\documentclass[onecolumn]{openhelix}

\usepackage[most]{tcolorbox}

\usepackage{amsmath}
\usepackage{amssymb}
\usepackage{bbm}
\usepackage{tabularx}
\usepackage{float}
\usepackage{utfsym}
\usepackage{array}
\usepackage{booktabs}
\usepackage{makecell}
\usepackage{xspace}
\usepackage{algorithm}
\usepackage{algpseudocode}
\usepackage{fontawesome5}

\newcommand{\huggingface}{\raisebox{-1.5pt}{\includegraphics[height=1.05em]{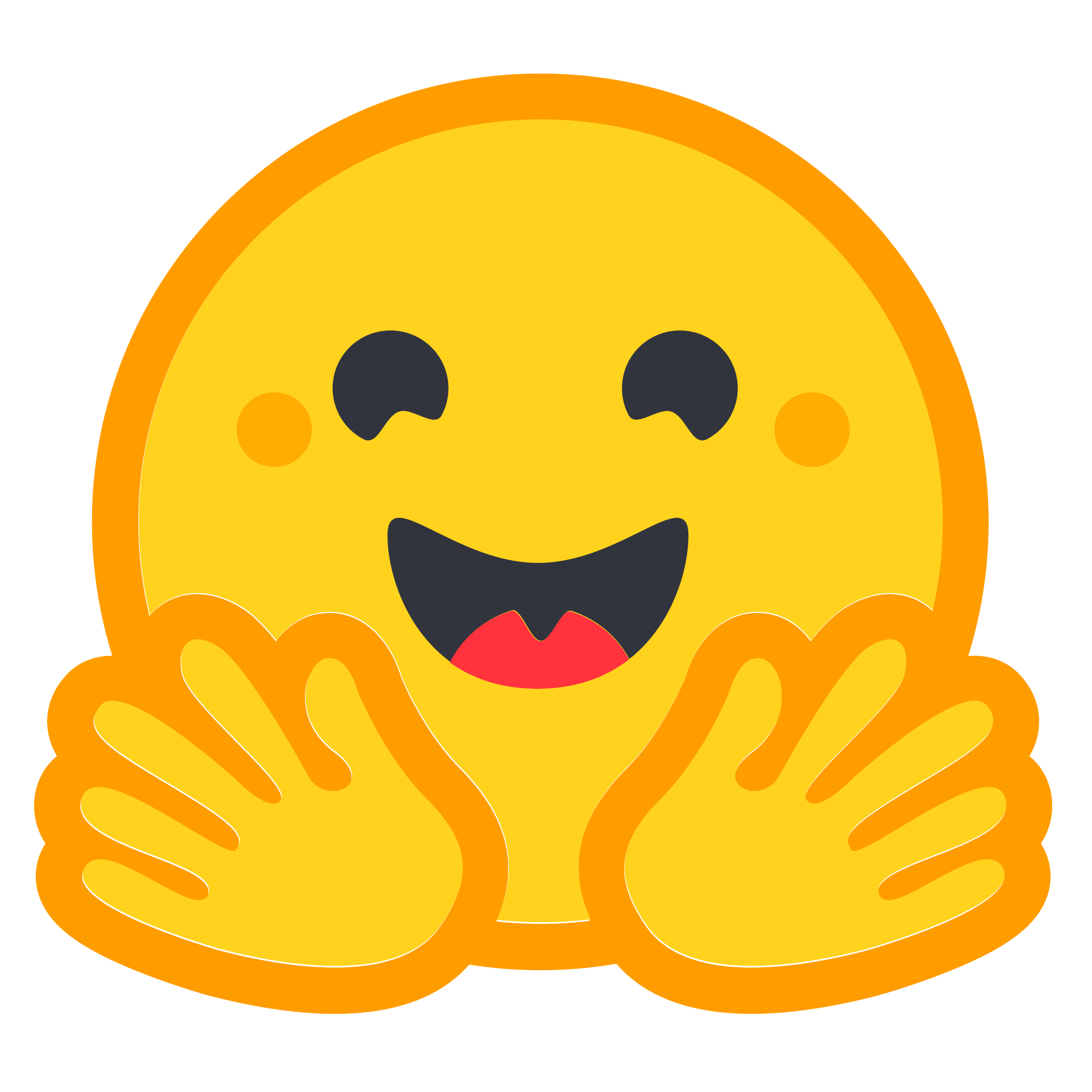}}\xspace}

\newcommand{\github}{\raisebox{-1.5pt}{\includegraphics[height=1.05em]{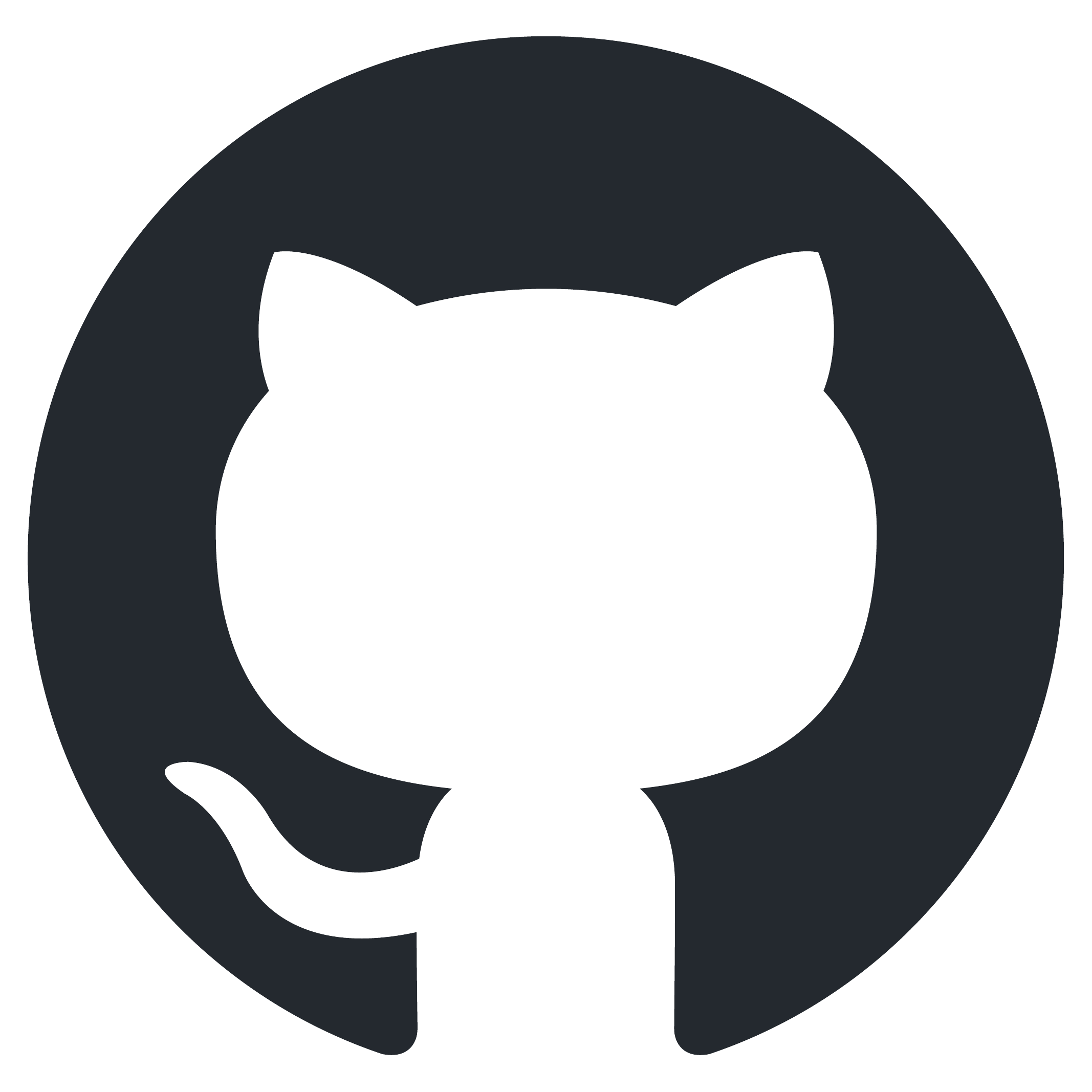}}\xspace}
\newcommand{\globe}{\raisebox{-1.5pt}{\faGlobe}\xspace}

\usepackage[table]{xcolor}
\definecolor{newyellow}{HTML}{FFD94D}
\definecolor{newgrey}{HTML}{7F7F7F}
\definecolor{newpink}{HTML}{FBCDF4}
\definecolor{realworldoft}{HTML}{029533}
\definecolor{realworldsf}{HTML}{8CC46A}
\definecolor{realworldour}{HTML}{FFC715}
\newcommand{\method}{\texttt{VAMPO}}
\newcommand{\methodname}{\method}

\title{VAMPO: Policy Optimization for Improving Visual Dynamics in Video Action Models}

\author[1,12\dagger]{Zirui Ge}
\author[1,2,12\dagger\ddagger]{Pengxiang Ding}
\author[4,12\dagger]{Baohua Yin}
\author[5,12]{Qishen Wang}
\author[6]{Zhiyong Xie}
\author[3,12]{Yemin Wang}
\author[13]{Jinbo Wang}
\author[7,12]{Hengtao Li}
\author[10,12]{Runze Suo}
\author[9,12]{Wenxuan Song}
\author[1,2,12]{Han Zhao}
\author[9,12]{Shangke Lyu}
\author[12]{Zhaoxin Fan}
\author[8]{Haoang Li}
\author[12\ddagger]{Ran Cheng}
\author[11]{Cheng Chi}
\author[1]{Huibin Ge}
\author[1*]{Yaozhi Luo}
\author[2*]{Donglin Wang}

\affiliation[1]{Zhejiang University}
\affiliation[2]{Westlake University}
\affiliation[3]{Xiamen University}
\affiliation[4]{University of Sussex}
\affiliation[5]{Tianjin University}
\affiliation[6]{Wuhan University}
\affiliation[7]{Hebei University of Technology}
\affiliation[8]{HKUST (GZ)}
\affiliation[9]{Nanjing University}
\affiliation[10]{Fudan University}
\affiliation[11]{Beijing Academy of Artificial Intelligence}
\affiliation[12]{OpenHelix Robotics}
\affiliation[13]{South China University of Technology}

\contribution[\dagger]{Equal Contributions}
\contribution[\ddagger]{Project Lead}
\contribution[*]{Corresponding Authors}

\abstract{

\begin{abstract}

Video action models are an appealing foundation for Vision--Language--Action systems because they can learn visual dynamics from large-scale video data and transfer this knowledge to downstream robot control. Yet current diffusion-based video predictors are trained with likelihood-surrogate objectives, which encourage globally plausible predictions without explicitly optimizing the precision-critical visual dynamics needed for manipulation. This objective mismatch often leads to subtle errors in object pose, spatial relations, and contact timing that can be amplified by downstream policies.
We propose \methodname{}, a post-training framework that directly improves visual dynamics in video action models through policy optimization. Our key idea is to formulate multi-step denoising as a sequential decision process and optimize the denoising policy with rewards defined over expert visual dynamics in latent space. To make this optimization practical, we introduce an Euler Hybrid sampler that injects stochasticity only at the first denoising step, enabling tractable low-variance policy-gradient estimation while preserving the coherence of the remaining denoising trajectory. We further combine this design with GRPO and a verifiable non-adversarial reward.
Across diverse simulated and real-world manipulation tasks, \methodname{} improves task-relevant visual dynamics, leading to better downstream action generation and stronger generalization.
\end{abstract}
}

\correspondence{%
\begin{tabular}[t]{@{}l@{\ }l@{}}
Pengxiang Ding at & \email{dingpx2015@gmail.com}
\end{tabular}%
}
\metadata[\globe\ Project Website]{\url{https://vampo-robot.github.io/VAMPO}}
\metadata[\huggingface\ Model]{\url{https://huggingface.co/williammmgezju}}
\metadata[\github\ Code]{\url{https://github.com/OpenHelix-Team/VAMPO}}

\begin{document}

\maketitle

\section{Introduction}

\begin{figure}[t]
    \centering
    \includegraphics[width=\linewidth]{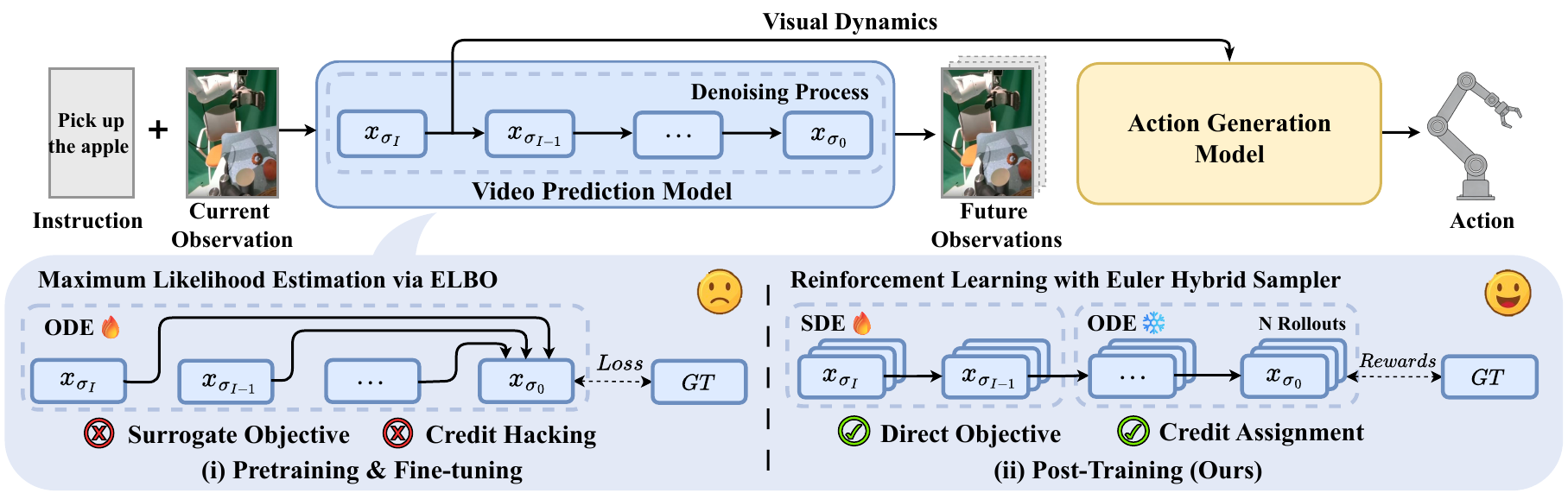}
    \caption{\textbf{Overall of \methodname.} Our post-training framework introduces reinforcement learning from verified rewards in place of the surrogate objective in video action models, enabling direct optimization of task-specific goals in training video prediction model (VPM). This approach improves the accuracy of VPM's predictive visual representations, leading to enhanced action generation and task performance. Notably, our method demonstrates significant improvements not only in simulated environments but also in real-world scenarios, showcasing its robustness and versatility across diverse settings.}
    \label{fig:teaser}
\end{figure}

Video action models are a promising foundation for Vision--Language--Action (VLA) systems because they can learn rich \textbf{visual dynamics} directly from video data \cite{hu2024video}. This provides two practical benefits. First, visual dynamics can be learned from large-scale video corpora without action annotations, enabling scalable pretraining. Second, the learned dynamics offer a strong prior for downstream action learning from limited robot demonstrations, since manipulation-relevant state evolution---such as pose changes and contact progression---is naturally reflected in videos.

A central challenge, however, is how to represent and expose such dynamics for downstream control. Early approaches often predict \textbf{pixel-level} future or goal images, which are intuitive but computationally expensive due to multi-step denoising, and often contain redundant appearance details that are only weakly relevant to action selection \cite{du2023learning, black2023zero, bu2024closed}. Consequently, recent work has increasingly shifted toward \textbf{latent-level} representations, which encode dynamics more compactly and can be extracted efficiently, for example from early denoising features in diffusion models \cite{wu2023unleashing, wen2024vidman, tian2024predictive, hu2024video, liu2025trivla} or through one-step distillation \cite{ye2026world}. These representations improve inference efficiency while placing greater emphasis on geometry- and relation-centric cues.

Despite this progress, existing pipelines still struggle to provide the precision-critical visual dynamics required for manipulation. Diffusion-based predictors are typically trained with likelihood-surrogate objectives, such as ELBO-style formulations \cite{kingma2021variational}, which prioritize global plausibility under the data distribution but do not explicitly optimize the accuracy of precision-sensitive factors, including object pose, fine-grained spatial relations, and contact timing \cite{black2023training}. As a result, the predicted representations may appear globally coherent while still containing subtle but important errors. When such errors are consumed by a downstream policy, they can be amplified near decision boundaries, leading to inaccurate actions and compounding failures over time.

To address this objective mismatch, we introduce \methodname, a method that improves visual dynamics in video action models through policy optimization (Figure~\ref{fig:teaser}). Our key idea is to view multi-step denoising as a sequential decision process \cite{black2023training}. Starting from a noisy latent, each denoising step takes an action---namely, the denoiser update---which induces a state transition defined by the sampler dynamics. The process terminates at a predicted future latent that represents the model's forecast. We then optimize this denoising policy to maximize a terminal reward that measures consistency with \textbf{expert visual dynamics}, obtained by encoding the ground-truth future sequence into the same latent space. In this way, \methodname{} explicitly optimizes dynamics signals that are not adequately captured by likelihood-based training alone.

\methodname{} realizes this idea through two design choices tailored to learning \textbf{improved visual dynamics}. First, we introduce an \textbf{Euler Hybrid sampler} that injects SDE-style stochasticity only at the first denoising step \cite{karras2022elucidating}. This design yields a tractable Gaussian transition for low-variance policy-gradient estimation via reparameterization, while preserving the temporal coherence of the remaining deterministic denoising trajectory and alleviating long-horizon credit assignment difficulties. Second, we adopt \textbf{GRPO} \cite{xue2025dancegrpo} with a verifiable, non-adversarial reward that combines $L_1$ distance and cosine similarity to directly align predicted visual dynamics with expert visual dynamics.
Unlike conventional ELBO-based training, which optimizes a surrogate objective, \methodname{} directly optimizes task-relevant prediction quality through reward-based post-training. As illustrated in Figure~\ref{fig:teaser}, this leads to more accurate visual dynamics for downstream action generation, ultimately improving manipulation performance and generalization across both simulated and real-world settings.

This paper makes the following contributions:
\begin{itemize}
    \item We show that a key limitation of existing video action models for manipulation lies in their training objective: ELBO-style likelihood surrogates optimize global plausibility, but not the precision-critical visual dynamics that determine downstream control quality.

    \item We present \methodname{}, a post-training framework that improves video action models by casting multi-step denoising as a sequential decision process and optimizing it with rewards defined over expert visual dynamics in latent space.


    \item We demonstrate substantial gains in both simulated and real-world manipulation, showing that improving task-relevant visual dynamics leads to better action generation and stronger generalization.
\end{itemize}

\section{Revisiting the Video Action Model}
\label{sec:video_action_model}

\subsection{Existing Paradigms.}
A typical video action model comprises two modules: a video prediction model (VPM) and an action generation model (AGM). The VPM anticipates a representation of the future conditioned on the current observation and a language instruction, while the AGM plays the role of an inverse-dynamics component that converts the predicted future representation—optionally together with additional conditioning signals—into an executable action sequence. 
A key design choice concerns the representation produced by the VPM, which leads to two dominant families of video action models:

\textbf{(1) Pixel-level approaches.}
In pixel-level formulations \cite{du2023learning, black2023zero, bu2024closed}, the predicted future takes the form of an explicit goal observation, such as a future image or a short video clip. 
This design is attractive for its interpretability, since the model’s intended outcome can be directly visualized. However, predicting in pixel space is often unnecessary for control: it forces the model to synthesize high-frequency appearance details (e.g., texture and illumination) that are only weakly coupled to action selection, introducing redundancy and potentially allowing prediction errors in visual details to propagate into downstream decision making.

\textbf{(2) Latent-level approaches.}
In latent-level formulations \cite{wu2023unleashing, wen2024vidman, tian2024predictive, hu2024video, liu2025trivla}, the VPM outputs a compact intermediate representation rather than decoded pixels, and the action generator conditions primarily on this representation. For diffusion-based VPMs in particular, the relevant future representation is commonly extracted from intermediate denoising features rather than from fully generated images. Such latent representations tend to suppress nuisance visual variability while retaining task-relevant dynamics and affordances, thereby enabling more efficient and robust control \cite{wu2023unleashing,wen2024vidman,tian2024predictive,hu2024video}. Moreover, intermediate denoising features can often be obtained using only the earliest denoising step(s), substantially reducing inference cost and mitigating the runtime bottleneck that would otherwise hinder real-time robotic deployment. Consequently, latent-level designs have become the prevailing choice in practice.
And in this work, we also adopt this paradigm and conduct our analyses under this setting.

\subsection{Instantiation of Latent-level Paradigm.}

Here, we follow the standard latent-level video action modeling paradigm~\cite{hu2024video} to instantiate our approach in a concrete form. Specifically, we consider a dataset $\mathcal{D} = \{(o, l, a, N)_i\}_{i=1}^M$ consisting of expert demonstrations, where each transition contains a current observation $o$, a language instruction $l$, a ground-truth action sequence $\mathbf{a}$, and a target future video clip $N$. Given a sample from $\mathcal{D}$, we form the video-model condition $c_v = (o,l)$. The VPM produces a predicted future representation $\hat{x}_0$ and an intermediate visual feature $\hat{h}$. Subsequently, the AGM generates an action sequence $\hat{a}_0$ conditioned on 
$c_a=(\hat{h},l)$:\begin{equation}(\hat{x}_0, \hat{h}) = \mathrm{VPM}_{\theta}(c_v), \qquad \hat{a}_0 = \mathrm{AGM}_{\theta}(c_a).\end{equation}

\textbf{Details of VPM.} We instantiate VPM with Stable Video Diffusion (SVD) \cite{blattmann2023stable}, a latent video diffusion model built on denoising diffusion models \cite{ho2020denoising} with EDM-style design and sampling \cite{karras2022elucidating}.
Let $N$ denote an expert demonstration video clip (a sequence of observations) aligned with condition $c_v$.
We encode $N$ into the latent space using a pretrained VAE encoder $\mathcal{E}$, yielding the clean latent $x_0 = \mathcal{E}(N)$, which serves as the target future representation.
Following EDM, the forward noising process directly adds Gaussian noise at a continuous noise level $\sigma$:
\begin{equation}
x_\sigma = x_0 + \sigma \epsilon, \quad \epsilon \sim \mathcal{N}(0, \mathbf{I}).
\end{equation}

The conditional denoiser $D_{\theta_v}(x_{\sigma_i}, \sigma_i, c_v)$ predicts the clean latent $x_0$ from a noisy sample $x_{\sigma_i}$ under guidance of $c_v$.
The supervised training objective is:
\begin{equation}
\mathcal{L}_{Video} = \mathbb{E}_{x_0,\,\epsilon,\,\sigma_i}\, \big\| D_{\theta_v}(x_{\sigma_i}, \sigma_i, c_v) - x_0 \big\|^2
\end{equation}

At inference time, given a decreasing noise schedule $\sigma_I > \cdots > \sigma_0 = 0$, we generate the predicted future representation by numerically integrating the probability-flow ODE with Euler Discrete solver:
\begin{equation}
x_{\sigma_{i-1}} = x_{\sigma_i} + (\sigma_{i-1} - \sigma_i)\, \frac{x_{\sigma_i} - D_{\theta_v}(x_{\sigma_i}, \sigma_i, c_v)}{\sigma_i},
\end{equation}

Following~\cite{hu2024video}, once the model is trained, we extract the predictive visual representation $\hat{h}$ from the multi-layer hidden states of $D_{\theta_v}$ at the \textbf{first} denoising step. This representation captures high-level spatiotemporal structure and task-relevant dynamics, and is used to condition the AGM.

\textbf{Details of AGM.} The AGM generates an action sequence $\mathbf{a}_0$ conditioned on $c_a=(\hat{h},l)$.
We adopt a diffusion policy \cite{chi2025diffusion} as the action head and use a DDIM-style denoising process \cite{song2020denoising}.
Given a ground-truth action sequence $\mathbf{a}_0$, the forward noising process constructs a noisy action at diffusion step $k$:
\begin{equation}
a_k = \sqrt{\bar{\beta}_k}\,a_0 + \sqrt{1-\bar{\beta}_k}\,\epsilon, \quad \epsilon \sim \mathcal{N}(0,\mathbf{I}),
\end{equation}
where $\bar{\beta}_k$ is the (cumulative) noise coefficient at step $k$.
The action denoiser $D_{\theta_a}$ is trained to reconstruct $\mathbf{a}_0$ by minimizing:
\begin{equation}
\mathcal{L}_{Action} = \mathbb{E}_{a_0,\,k}\, \big\| D_{\theta_a}(a_k, k, c_a) - a_0 \big\|^2.
\end{equation}

\subsection{Analysis of Current Paradigm.}
\textbf{Limitation.}
Diffusion-based VPMs are typically optimized with distribution-level objectives, such as likelihood surrogates, rather than direct objectives that enforce high-precision, control-relevant future representations. This leads to an \emph{objective mismatch}: improving likelihood does not necessarily produce representations that are reliable for downstream action generation.
In practice, video diffusion models are trained to approximately maximize data likelihood, most commonly through variational objectives such as the ELBO or related surrogate losses \cite{kingma2021variational,ho2020denoising}. This training paradigm introduces approximation at multiple levels: the optimized objective is itself only a proxy for true likelihood, and the finite-step stochastic sampling procedure may introduce additional error. As a result, the generated future representations are not explicitly encouraged to be accurate in the state variables that matter most for control.
Consequently, VPM outputs may be globally coherent while remaining subtly incorrect in either pixel space or latent space, with small errors in object pose, contact timing, or fine-grained spatial relations (e.g., slight misalignment, small gaps, or off-by-one-frame contact events). When such representations are used to condition an AGM, these discrepancies can be amplified. Near decision boundaries, even minor representational errors may alter action selection---for example, whether to close the gripper, which side of an obstacle to pass, or when to initiate contact. Once an incorrect action is executed, the environment state may deviate further from the intended trajectory, leading to compounding errors over time and ultimately lower task success rates.

\textbf{Solution.}
A natural way to address this objective mismatch is to directly optimize the VPM for the accuracy of its predicted visual dynamics, rather than relying solely on likelihood-based surrogate objectives. Concretely, we treat the multi-step denoising process as a stochastic sequential generator, define a reward by comparing the final prediction with a task-relevant target (e.g., ground-truth representations or oracle-derived features), and optimize the model using policy-gradient-style updates \cite{black2023training}. This objective explicitly penalizes subtle but control-critical errors in pose, contact, and spatial relations, thereby narrowing the gap between generative training and action-centric deployment. In the following sections, we describe how to instantiate this framework in practice.

\begin{figure*}[t]
    \centering
    \includegraphics[width=\linewidth]{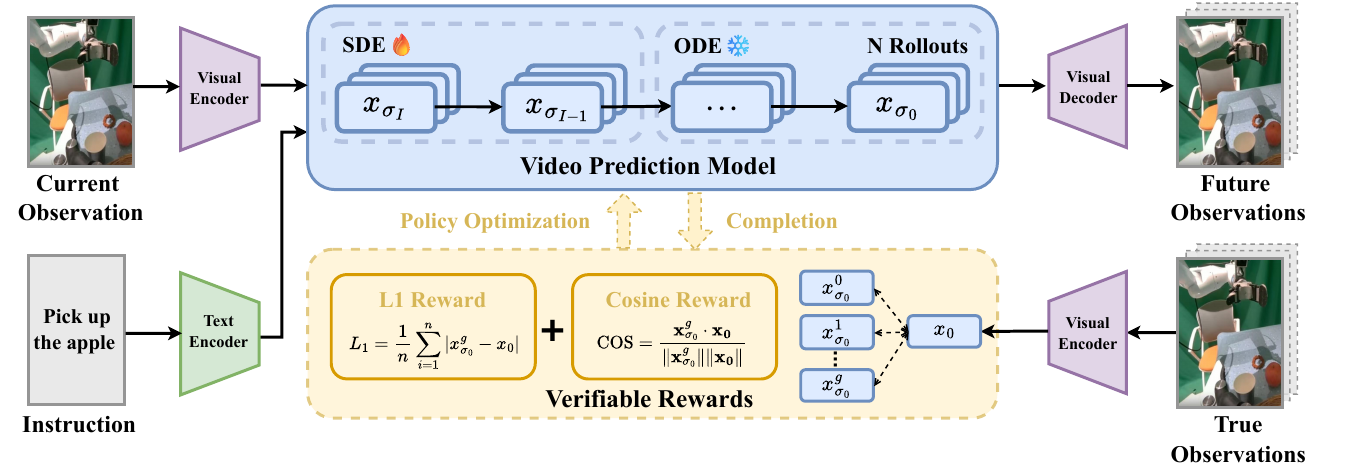}
    \caption{\textbf{Overview of the \methodname{} training paradigm.} In the pretraining stage, the video prediction model (VPM) and action generation model (AGM) are trained on expert demonstrations. In the policy optimization stage, the VPM generates future latents via a hybrid denoising process, using SDE-style stochasticity only at the first step and ODE-based denoising for the remaining steps. Verified rewards are computed by comparing predicted latents with expert latents, and GRPO is used to optimize the VPM toward more precise, control-relevant visual dynamics for downstream action generation.}
    \label{fig:overview}
\end{figure*}
\section{Method}


\subsection{Problem formulation}
\label{sec:rlvr_vpm}

Following DDPO \cite{black2023training}, we model the denoising process of a diffusion-based VPM as an MDP $\mathcal{M} = (\mathcal{S}, \mathcal{A}, \boldsymbol{\rho}_0, \boldsymbol{P}, \boldsymbol{R})$, which consists of five key components: the state space $\mathcal{S}$, action space $\mathcal{A}$, initial state distribution $\boldsymbol{\rho}_0$, transition kernel $\boldsymbol{P}$, and reward function $\boldsymbol{R}$. For a denoising trajectory of length $I$,
\begin{equation}
\tau = (\boldsymbol{s}_I, \boldsymbol{a}_I, \boldsymbol{s}_{I-1}, \boldsymbol{a}_{I-1}, \dots, \boldsymbol{s}_0, \boldsymbol{a}_0),
\end{equation}
where for each denoising step $i \in \{0, \dots, I\}$, the state $\boldsymbol{s}_i \in \mathcal{S}$ and action $\boldsymbol{a}_i \in \mathcal{A}$ are determined by the denoiser policy $\boldsymbol{\pi}(\boldsymbol{a} \mid \boldsymbol{s})$. Specifically, the components are defined as follows:
\begin{equation}
\begin{aligned}
\boldsymbol{s}_i &\triangleq (x_{\sigma_i}, \sigma_i, c_v), \\
\boldsymbol{a}_i &\triangleq x_{\sigma_{i-1}}, \\
\boldsymbol{\rho}_0(\boldsymbol{s}_I) &\triangleq (p(c_v), \delta_I, \epsilon), \\
\boldsymbol{P}(\boldsymbol{s}_{i-1} \mid \boldsymbol{s}_i, \boldsymbol{a}_i)
&\triangleq (\delta_{x_{\sigma_{i-1}}}, \delta_{\sigma_{i-1}}, \delta_c), \\
\boldsymbol{R}(\boldsymbol{s}_i, \boldsymbol{a}_i) &\triangleq \mathbbm{1}_{[i=0]} \cdot r(x_{\sigma_0}), \\
\boldsymbol{\pi}(\boldsymbol{a}_i \mid \boldsymbol{s}_i) &\triangleq p(x_{\sigma_{i-1}} \mid x_{\sigma_i}, c_v).
\end{aligned}
\end{equation}

Here, $\delta$ is the Dirac delta distribution. The initial state distribution $\boldsymbol{\rho}_0$ is determined by the condition distribution $p(c_v)$, starting with timestep $I$ and Gaussian white noise $\epsilon$. The reward function is only provided at the final denoising step $i=0$, where $r(x_{\sigma_0})$ is the reward function used to evaluate the quality of the generated output. The transition kernel $\boldsymbol{P}$ describes the deterministic state transitions after an action is taken.

Thus, the problem becomes a policy optimization problem where we aim to learn a policy that maximizes the reward by optimizing the model's prediction of the visual representation over time. Formally, we define our objective as
\begin{equation}
\mathcal{J}(\boldsymbol{\pi}) = \mathbb{E}_{\tau \sim \boldsymbol{\pi}} \left[ \sum_{i=0}^{I} \boldsymbol{R}(\boldsymbol{s}_i, \boldsymbol{a}_i) \right].
\end{equation}

\subsection{Denoising with SDE-Sampling}
\label{sec:sde_sampling}

In the original VPM pipeline, denoising is performed via a deterministic policy, denoted as Euler Discrete solver, consistent with the deterministic probability-flow ODE framework \cite{song2020score}. This formulation yields a deterministic denoising path rooted in the initial state.
Since this deterministic process provides no stochastic exploration and lacks the tractable transition density $p(x_{\sigma_{i-1}}\mid x_{\sigma_i},c_v)$ required by policy gradient methods \cite{black2023training}, we instead employ the Euler-Ancestral sampler \cite{karras2022elucidating}. This approach transforms the ODE update into an SDE transition through step-wise Gaussian noise injection.
For a transition from $\sigma_i$ to $\sigma_{i-1}$, we sample:
\begin{equation}
x_{\sigma_{i-1}} = x_{\sigma_{i-1}}^{\text{det}} + \sigma_{\text{up}} \epsilon, \quad \epsilon \sim \mathcal{N}(0, \mathbf{I}).
\end{equation}
Here, the deterministic component is
\begin{equation}
x_{\sigma_{i-1}}^{\text{det}} = x_{\sigma_i} + (\sigma_{\text{down}} - \sigma_i)\, \frac{x_{\sigma_i} - D_{\theta_v}(x_{\sigma_i}, \sigma_i, c_v)}{\sigma_i},
\end{equation}
where $D_{\theta_v}$ is the VPM denoiser network, and the noise level is decomposed into deterministic and stochastic parts:
\begin{align}
\sigma_{\text{up}} &= \sqrt{\sigma_{i-1}^2 \frac{\sigma_i^2 - \sigma_{i-1}^2}{\sigma_i^2}}, \\
\sigma_{\text{down}} &= \sqrt{\sigma_{i-1}^2 - \sigma_{\text{up}}^2}.
\end{align}
This yields an explicit Gaussian transition density:
\begin{equation}
p_\theta(x_{\sigma_{i-1}} \mid x_{\sigma_i}, c_v) \;=\; \mathcal{N}\!\left(x_{\sigma_{i-1}};\, x_{\sigma_{i-1}}^{\text{det}},\, \sigma_{\text{up}}^2 \mathbf{I}\right).
\end{equation}
We denote $\boldsymbol{\pi}^{eas}(\boldsymbol{a}\mid \boldsymbol{s})$ as the denoiser policy with the Euler Ancestral sampler. 

\subsection{Reinforcement Learning with Hybrid Sampler}
\label{sec:rlvr_grpo}

In our framework, we employ GRPO \cite{xue2025dancegrpo} to fine-tune the VPM, enabling it to better capture the complex dynamics of future observations. Our methodology is structured into two primary stages: Rollout with Euler Hybrid sampler and Optimization with verifiable reward.

\noindent\textbf{Rollout with Euler Hybrid sampler.}
For a given conditional input $c_v$, the VPM must produce a group of $G$ diverse candidate future representations $\{ x^g \}_{g=1}^G$, which is crucial for stable policy optimization. Specifically, we replace the Euler Discrete solver with the Euler Ancestral sampler $\boldsymbol{\pi}^{eas}$, transforming the denoising process from an ODE formulation into an SDE-based sampling procedure. However, introducing stochasticity at all denoising steps leads to severe credit assignment issues: since the entire trajectory shares the same reward signal, the policy may maximize the return by exploiting later denoising actions that are irrelevant to action modeling, rather than genuinely improving the early-step visual representation that is critical for downstream action decision making. 

Therefore, we design the Euler Hybrid sampler, which applies SDE sampling only at the first denoising step before the action-relevant representation is captured and keeps the remaining steps deterministic:
\begin{equation}
p(\boldsymbol{s}_{0:I-1}, \boldsymbol{a}_{1:I} \mid \boldsymbol{s}_I)
=
\boldsymbol{\pi}^{eas}(\boldsymbol{a}_I \mid \boldsymbol{s}_I)
\prod_{i=1}^{I-1}
\delta\!\left(
\boldsymbol{a}_i - f_{\mathrm{eds}}(\boldsymbol{s}_i)
\right),
\end{equation}

where $f_{\mathrm{eds}}$ denotes the deterministic Euler Discrete update function that maps the current denoising state to the next latent via a single Euler ODE step. Moreover, this design further improves training stability and efficiency, as limiting stochasticity to a single denoising step preserves the pre-trained temporal consistency of the video model and reduces the number of steps involved in backpropagation. The full training procedure is summarized in Algorithm~\ref{alg:grpo}.

\begin{algorithm}[t]
\caption{\textbf{GRPO Training for Video Prediction Model.}}
\label{alg:grpo}
\begin{algorithmic}[1]
\Require Initial policy $\boldsymbol{\pi}_{\theta_{\text{old}}}$ , dataset $\mathcal{D}$, group size $G$
\Ensure Optimized policy $\boldsymbol{\pi}_\theta$
\State Initialize $\boldsymbol{\pi}_\theta \leftarrow \boldsymbol{\pi}_{\theta_{\text{old}}}$
\For{iteration $t = 1, 2, \ldots, T$}
    \State Sample batch of conditions $c_v=(o,l)$ and expert future representations $x_0$ from $\mathcal{D}$
    \State Sample shared initial noise $x_{\sigma_I} \sim \mathcal{N}(0,\sigma_I^2\mathbf{I})$
    \For{$g = 1$ to $G$}
        \State Roll out hybrid sampling: first step SDE, remaining steps ODE $\rightarrow$ obtain $x_{\sigma_0}^g$
        \State Compute reward $r_g$ from $(x_{\sigma_0}^g, x_0)$
    \EndFor
    \State Compute advantages $A_g$ via group normalization
    \State Update policy parameters $\theta$ by maximizing $J(\theta)$
    \State Update $\boldsymbol{\pi}_{\theta_{\text{old}}} \leftarrow \boldsymbol{\pi}_\theta$ periodically
\EndFor
\end{algorithmic}
\end{algorithm}

\noindent\textbf{Optimization with verifiable reward.}
Following the stochastic rollouts, which provide a diverse set of candidate samples $g \in \{1, \dots, G\}$ ending in a denoised latent $x_{\sigma_0}^g$, we evaluate the predicted future representation against the ground truth. Let $x_0$ be the expert future representation obtained by encoding the corresponding expert future clip with the VAE encoder.
For each sampled representation, we compute a verifiable relative latent-consistency reward:
\begin{equation}
r = -\lambda_{L1}\,\|x_{\sigma_0}-x_0\|_1 + \lambda_{\text{cos}}\,\frac{x_{\sigma_0}\cdot x_0}{\|x_{\sigma_0}\|\,\|x_0\|}.
\end{equation}
We then compute group-normalized advantages:
\begin{equation}
A_g = \frac{r_g - \text{mean}(\{r_1, r_2, \dots, r_G\})}{\text{std}(\{r_1, r_2, \dots, r_G\})}.
\end{equation}
Since the reward is terminal, we apply the same advantage $A_g$ to all stochastic denoising steps within the trajectory.

The GRPO objective is:
\begin{align}
J(\theta)
&= \mathbb{E}_{\boldsymbol{a}_{I,g} \sim \boldsymbol{\pi}_{\theta_\text{old}}(\cdot \mid \boldsymbol{s}_{I,g})}
\Big[
\frac{1}{G} \sum_{g=1}^{G} \min(
\rho_{I,g} A_g, 
\text{clip}(\rho_{I,g}, 1-\epsilon_c, 1+\epsilon_c) A_g
)
\Big]
\end{align}

\begin{equation}
\rho_{I,g} = \frac{\boldsymbol{\pi}_\theta(\boldsymbol{a}_{I,g} \mid \boldsymbol{s}_{I,g})}{\boldsymbol{\pi}_{\theta_\text{old}}(\boldsymbol{a}_{I,g} \mid \boldsymbol{s}_{I,g})}
\end{equation}

where  $\epsilon_{c}$ is the clipping threshold that prevents excessively large policy updates \cite{schulman2017proximal}. By maximizing $J(\theta)$, the VPM is optimized to align its generative distribution with high-reward denoising trajectories that stay consistent with expert dynamics.

\section{Experiments}
In this section, we aim to answer the following questions:

\textbf{Q1.} To what extent can our approach improve the quality of visual dynamics modeling?\\
\textbf{Q2.} Do improvements in visual dynamics modeling lead to measurable gains in policy performance?\\
\textbf{Q3.} Through what mechanisms do improved visual dynamics translate into better policy performance?\\
\textbf{Q4.} How does \methodname{} compare with state-of-the-art methods?\\
\textbf{Q5.} What is the contribution of each core component to the overall performance?

\subsection{Evaluation on Simulation Environment}

\textbf{Simulation Settings.}

We evaluate on the CALVIN benchmark~\cite{mees2022calvin} for long-horizon, language-conditioned robotic manipulation. Following the ABC$\rightarrow$D protocol, models are trained on environments ABC and evaluated on the unseen environment D, which differs in visual appearance and layout, requiring both long-horizon execution and generalization. We further assess long-horizon capability on L-CALVIN~\cite{fan2025long}, which extends task sequences from 5 to 10 steps for more challenging multi-step evaluation.

\textbf{Baseline.}
We adopt~\cite{hu2024video} as our base policy as it is a highly competitive VPM-based VLA with strong visual dynamics modeling capability.
Beyond~\cite{hu2024video}, we compare against representative methods from both VLM-based and VPM-based VLAs. For VLM-based VLAs, we include~\cite{black2024pi_0,intelligence2025pi_}; for VPM-based VLAs, we evaluate both pixel-level and latent-level variants such as~\cite{tian2024predictive}. Moreover, to ensure a fair and comprehensive assessment, we also report results of existing techniques that further strengthen our base policy, highlighting a key advantage of our approach: it improves capability without requiring any architectural modifications.

\textbf{Training Details.}
We perform post-training on a VPM initialized from the pretrained checkpoint released by VPP, using videos from the CALVIN ABC dataset, which contains 18,033 trajectories. From a selected subset of trajectories, we split the data into 129,454 video samples for training. The VPM is trained around 1.5k steps on 64 NVIDIA H20 GPUs. Then, we train the AGM on the whole Calvin dataset for approximately 10 epochs using 8 NVIDIA H20 GPUs. Evaluations are conducted on NVIDIA RTX 5880 GPUs.

\begin{figure*}[htbt]
    \centering
    \includegraphics[
    width=\linewidth,
    trim=0cm 0cm 0cm 0cm,
    clip
]{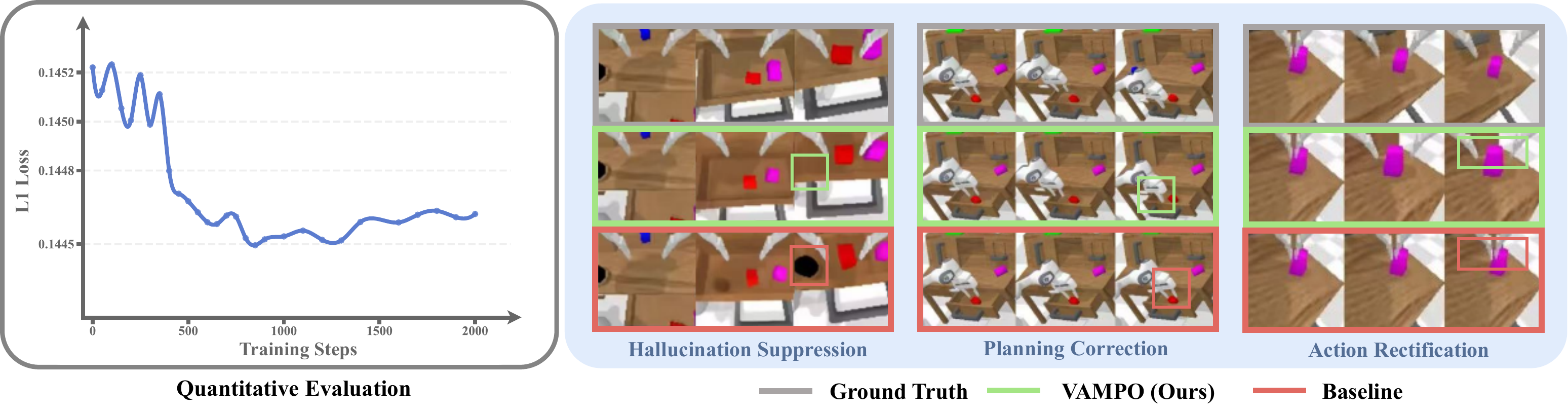}
    \caption{\textbf{Evaluation on Visual Dynamics.} The figure reports the L1 evaluation between predicted latents and ground-truth latents over training steps, and \methodname{} exhibits improved alignment with expert dynamics, leading to hallucination suppression, planning correction, and action rectification.}
    \label{fig:fig3}
\end{figure*}

\textbf{Analysis of Visual Dynamics.}
As shown in Figure~\ref{fig:fig3}, the quality of the learned latent representations, measured in terms of visual dynamics, exhibits an overall improving trend throughout training. Although fluctuations are observed at intermediate stages, the visual dynamics encoded in the latent space become progressively more coherent and structured as training proceeds.
In addition, a qualitative comparison between the baseline and predicted future observations demonstrates improved alignment with expert dynamics. Specifically, the optimized VPM produces latent representations that decode into more accurate object poses, spatial relationships, and contact progression, which in turn facilitate more effective action generation.

\textbf{Improvement in Action Execution.}
As shown in Table~\ref{tab:effectiveness-better-visual-dynamics}, even when the downstream AGM is frozen, optimizing the VPM consistently improves overall performance. Although the gains are modest, this is expected because the downstream policy is not co-optimized.
Furthermore, after fine-tuning the VPM with our method, enabling training of the downstream AGM leads to substantially larger improvements. These results indicate that unoptimized visual dynamics can misguide downstream policy learning (equivalently, the mapping learned by the inverse dynamics model). Once the visual dynamics are properly optimized, more accurate visual-dynamics representations can be translated into better action dynamics, highlighting the advantage of our approach that directly updates visual dynamics via policy-gradient optimization.

\textbf{Correlation between Visual Dynamics and Action}
To characterize the coupling between visual dynamics and actions, we further measure the \textbf{effective rank} (ER) of the Jacobian $\mathbf{J}=\partial \mathbf{a}/\partial \mathbf{v}$ \cite{roy2007effective}. After singular value decomposition (SVD) with singular values $\{\sigma_i\}$, ER is defined as
\[
\mathrm{ER}(\mathbf{J})=\frac{\left(\sum_i \sigma_i\right)^2}{\sum_i \sigma_i^{2}}.
\]
We report the \textbf{Average ER} and the normalized \textbf{Average ER Ratio (ERR)} $=\overline{\mathrm{ER}}/\mathrm{ER}_{\max}$, with $\mathrm{ER}_{\max}\le\min(d_a,d_v)$. Larger ER indicates richer vision--action coupling.
As shown in Table~\ref{tab:effectiveness-better-visual-dynamics}, compared with the base policy, our proposed paradigm achieves substantial improvements in both Avg.\ ER and Avg.\ ERR. The SVD analysis reveals that \methodname{} exhibits a more gradual cumulative contribution curve and maintains larger singular values across a broader range of indices, indicating that the downstream AGM relies on a larger number of mutually independent visual directions in the learned visual dynamics when generating actions—i.e., the vision--action coupling becomes richer—which in turn leads to fundamental performance gains during the action execution stage.

\begin{table*}[!ht]
    \centering
    \small
    \setlength{\tabcolsep}{2.5pt}
    \caption{\textbf{Effectiveness of improved visual dynamics on CALVIN ABC$\to$D.} 
    The table reports task completion in a row (1--5), average trajectory length (Avg.\ Len), 
    and vision--action coupling metrics (Avg.\ ER, Avg.\ ERR). Post-training both VPM and AGM with 
    \methodname{} achieves the best performance; improvements over the base policy are in \textbf{bold}.}
    \label{tab:effectiveness-better-visual-dynamics}
    
    \begin{tabular}{cc|ccccc|c|cc}
        \toprule
        \multicolumn{2}{c|}{\multirow{2}{*}{CALVIN ABC$\to$D}} 
        & \multicolumn{5}{c|}{Task completed in a row $\uparrow$} 
        & Avg.\ Len $\uparrow$ 
        & Avg.\ ER $\uparrow$ 
        & Avg.\ ERR $\uparrow$ \\
        
        \cmidrule(lr){3-7}
        \multicolumn{2}{c|}{} & 1 & 2 & 3 & 4 & 5 & & & \\
        \midrule
        
        \multirow{1}{*}{} 
        & Base Policy 
        & 96.0 & 91.3 & 86.4 & 80.4 & 74.7 
        & 4.28 
        & 29.28 
        & 0.0603 \\
        
        \midrule
        
        \multirow{1}{*}{+} 
        & post-training VPM + original AGM 
        & 96.3 & 91.4 & 87.0 & 82.9 & 77.2 
        & 4.35 
        & -- 
        & -- \\
        
        \midrule
        
        \multirow{1}{*}{+} 
        & post-training VPM + post-training AGM 
        & \textbf{98.0} & \textbf{94.8} & \textbf{91.3} & \textbf{88.3} & \textbf{83.1} 
        & \textbf{4.56} 
        & \textbf{43.88} 
        & \textbf{0.0814} \\
        
        \bottomrule
    \end{tabular}
\end{table*}

\begin{table*}[!h]
    \centering
    \small
	\setlength{\tabcolsep}{2pt}
	\caption{\textbf{Performance on the CALVIN ABC$\to$D benchmark.} The table reports task completion in a row (1--5) and average trajectory length (Avg.\ Len) for VLM-based and VPM-based VLAs. \methodname (Ours) achieves the best performance; our method's row is highlighted.}
    \label{tab:calvin-benchmark}
	\begin{tabular}{cc|ccccc|c}
		\toprule
		\multicolumn{2}{c|}{\multirow{2}{*}{CALVIN ABC$\to$D}} & \multicolumn{5}{c|}{Task completed in a row $\uparrow$} & Avg.\ Len $\uparrow$ \\
		\cmidrule(lr){3-7}
		\multicolumn{2}{c|}{} & 1 & 2 & 3 & 4 & 5 & \\
		\midrule
		\multirow{4}{*}{\textbf{VLM-based VLA}}
		& OpenVLA \citep{kim2024openvla}\textsubscript{\textbf{(CoRL)}} & 91.3 & 77.8 & 62.0 & 52.1 & 43.5 & 3.27 \\
		& OpenVLA-OFT \citep{kim2025fine}\textsubscript{\textbf{(RSS)}} & 96.3 & 89.1 & 82.4 & 75.8 & 66.5 & 4.10 \\
		& $\pi_0$ \citep{black2024pi_0}\textsubscript{\textbf{(RSS)}} & 93.7 & 83.2 & 74.0 & 62.9 & 51.0 & 3.65 \\
		& $\pi_{0.5}$ \citep{intelligence2025pi_}\textsubscript{\textbf{(CoRL)}} & 92.7 & 84.3 & 76.7 & 68.8 & 61.3 & 3.84 \\
		\midrule
		\multirow{4}{*}{\textbf{VPM-based VLA (pixel-level)}}
		& UniPi \citep{du2023learning}\textsubscript{\textbf{(NeurIPS)}} & 56.0 & 16.0 & 8.0 & 8.0 & 4.0 & 0.92 \\
		& SuSIE \citep{black2023zero}\textsubscript{\textbf{(ICLR)}} & 87.0 & 69.0 & 49.0 & 38.0 & 26.0 & 2.69 \\
		& CLOVER \citep{bu2024closed}\textsubscript{\textbf{(NeurIPS)}} & 96.0 & 83.5 & 70.8 & 57.5 & 45.4 & 3.53 \\
		\midrule
		\multirow{6}{*}{\textbf{VPM-based VLA (latent-level)}}
		& GR-1 \citep{wu2023unleashing}\textsubscript{\textbf{(ICLR)}} & 85.4 & 71.2 & 59.6 & 49.7 & 40.1 & 3.06 \\
		& VidMan \citep{wen2024vidman}\textsubscript{\textbf{(NeurIPS)}} & 91.5 & 76.4 & 68.2 & 59.2 & 46.7 & 3.42 \\
		& Seer \citep{tian2024predictive}\textsubscript{\textbf{(ICLR)}} & 94.4 & 87.2 & 79.9 & 72.2 & 64.3 & 3.98 \\
		& Seer\textsubscript{\textbf{Large}}\citep{wen2024vidman}\textsubscript{\textbf{(ICLR)}} & 96.3 & 91.6 & 86.1 & 80.3 & 74.0 & 4.28 \\
		& VPP \citep{hu2024video}\textsubscript{\textbf{(ICML)}} & 96.0 & 91.3 & 86.4 & 80.4 & 74.7 & 4.28 \\
		& Tri-VLA \citep{liu2025trivla}\textsubscript{\textbf{(Arxiv)}} & 96.8 & 92.4 & 86.8 & 83.2 & 81.8 & 4.37 \\
		\midrule
		\rowcolor[rgb]{.92,.92,.92}
		& \textbf{\methodname (Ours)} & \textbf{98.0} & \textbf{94.8} & \textbf{91.3} & \textbf{88.3} & \textbf{83.1} & \textbf{4.56} \\
		\bottomrule
	\end{tabular}%
\end{table*}%

\begin{table*}[!t]
    \centering
    \small
    \setlength{\tabcolsep}{1.5pt}

    \newsavebox{\tempbox}
    \savebox{\tempbox}{
        \begin{tabular}{cc|cccccccccc|c}
            \toprule
            Train$\rightarrow$Test & Method & \multicolumn{10}{c|}{Tasks completed in sequence $\uparrow$} & Avg.\ Len $\uparrow$ \\
            \cmidrule(lr){3-12}
            & & 1 & 2 & 3 & 4 & 5 & 6 & 7 & 8 & 9 & 10 & \\
            \midrule
            \multirow{4}{*}{ABC$\to$D}
            & OpenVLA \citep{kim2024openvla} & 0.67 & 0.34 & 0.24 & 0.12 & 0.03 & 0.01 & 0.01 & 0.01 & 0.00 & 0.00 & 1.43 \\
            & $\pi_0$ \citep{black2024pi_0} & 0.84 & 0.64 & 0.51 & 0.43 & 0.34 & 0.25 & 0.21 & 0.17 & 0.11 & 0.11 & 3.61 \\
            & VPP \citep{hu2024video} & 0.94 & 0.82 & 0.75 & 0.63 & 0.53 & 0.49 & 0.43 & 0.35 & 0.31 & 0.28 & 5.53 \\
            \midrule
            \rowcolor[rgb]{.92,.92,.92}
            & \textbf{\methodname (Ours)} & \textbf{0.97} & \textbf{0.89} & \textbf{0.81} & \textbf{0.75} & \textbf{0.71} & \textbf{0.64} & \textbf{0.61} & \textbf{0.51} & \textbf{0.45} & \textbf{0.39} & \textbf{6.73} \\
            \bottomrule
        \end{tabular}
    }

    \caption{\textbf{Performance on L-CALVIN (long-horizon).} The table reports tasks completed in sequence (1--10) and average trajectory length (Avg.\ Len) under the ABC$\to$D protocol. \methodname{} (Ours) yields the best results across all task lengths; improvements are highlighted.}
    \label{tab:l-calvin}
    
    \usebox{\tempbox}

    \vspace{1.0em}

    \includegraphics[width=\wd\tempbox, trim=0cm 0cm 0cm 0cm, clip]{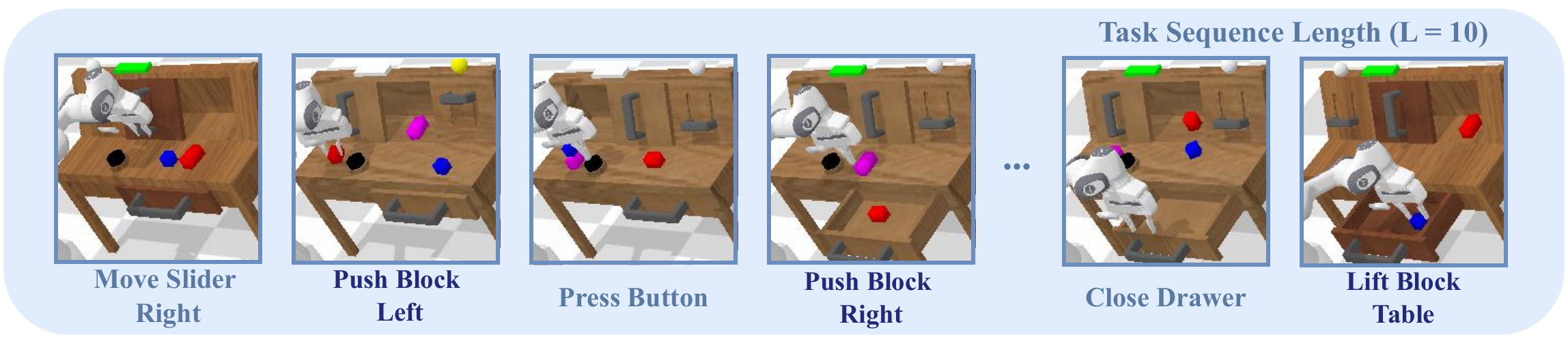}
    
    \makeatletter
    \def\@captype{figure}
    \makeatother
    \label{fig:l-calvin}
    
\end{table*}

\textbf{Comparison with SOTA Methods.} 
Table~\ref{tab:calvin-benchmark} and Table~\ref{tab:l-calvin} summarize comparisons with VLAs on CALVIN ABC$\to$D and L-CALVIN, respectively. We find that, although our method is built upon Video Prediction Policy (VPP)~\cite{hu2024video}, our post-training procedure—which directly optimizes visual dynamics without any architectural modifications—substantially improves VPP’s performance and also surpasses approaches based on other model families. Moreover, compared with prior VPP-based extensions such as~\cite{liu2025trivla}, our method requires neither additional network components nor extra data, yet seamlessly boosts the base model’s performance. Notably, on longer-horizon tasks in L-CALVIN (Table~\ref{tab:l-calvin}), \methodname{} yields larger gains, further highlighting the importance of improving visual dynamics modeling for long-horizon manipulation, where tasks involve richer and more diverse dynamics over extended time spans.

\subsection{Evaluation on Real-World}

\textbf{Real-World Settings.} To validate the effectiveness of our method in real-world scenarios, we evaluate video generation performance on three self-collected datasets that vary in robot embodiment and background conditions, and further assess downstream action execution on the Agibot Genie 01 platform. Details of the real-world action execution setup are illustrated in Figure~\ref{fig:real}.
All real-world experiments are conducted on the Agibot Genie 01 dual-arm robotic system, which features 14 DoF across two arms and a 2-DoF gripper on each end-effector, enabling flexible and coordinated manipulation. The platform is designed for tabletop daily manipulation tasks, including single-arm grasping, dual-arm coordination, and multi-object interaction.
The system is equipped with three RGB-D cameras for multi-view perception. A head-mounted Intel RealSense D455 provides a third-person global view of the tabletop layout, object distribution, and dual-arm motion states. In addition, two wrist-mounted Intel RealSense D405 cameras provide first-person close-up views, capturing fine-grained geometric details of grasp regions and contact interactions. All visual observations are synchronized with robot proprioceptive states and executed actions.
For each task, we collect 200 high-quality teleoperated demonstration trajectories in real-world environments. Each trajectory contains multi-view visual observations from the head and wrist cameras, robot joint states, end-effector poses, and executed actions.

\medskip
\noindent\textbf{Training Details.}
Training is conducted in two stages. In Stage 1, the model is trained for 1.5k steps to learn basic visual-semantic alignment and preliminary action generation strategies. In Stage 2, training continues for 10k additional steps to improve manipulation precision, coordination, and robustness.

\begin{figure*}[t] 
    \centering
    \includegraphics[width=\textwidth, trim=0 40 0 0, clip]{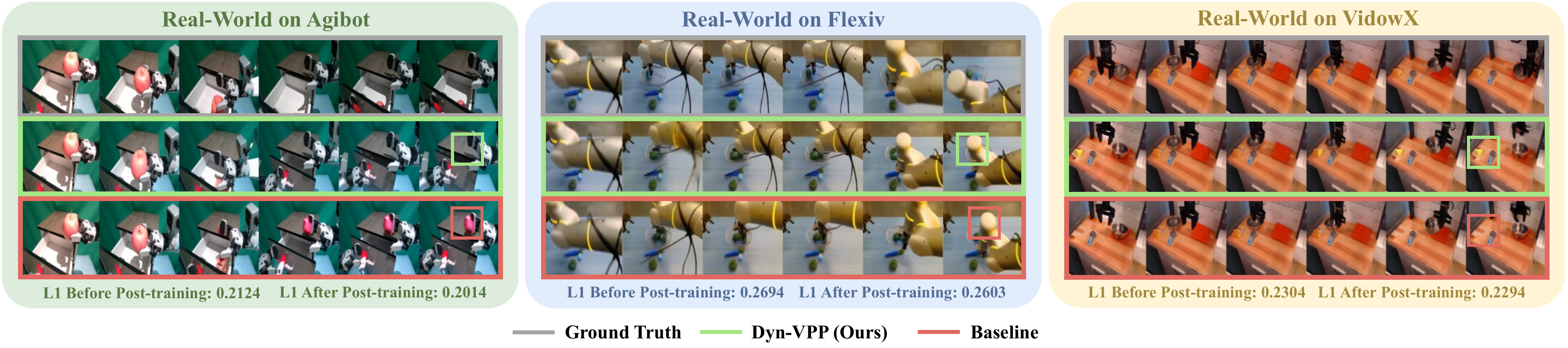}
    \caption{\textbf{Qualitative visualization across multiple benchmarks.} The figure presents improved action prediction and manipulation quality achieved by \methodname{} on real-world platforms (Agibot Genie 01, Flexiv dual-arm robot, VidowX).}
    \label{fig:vis_appendix}
\end{figure*}

\begin{figure*}[t]
    \centering
    \includegraphics[width=\linewidth]{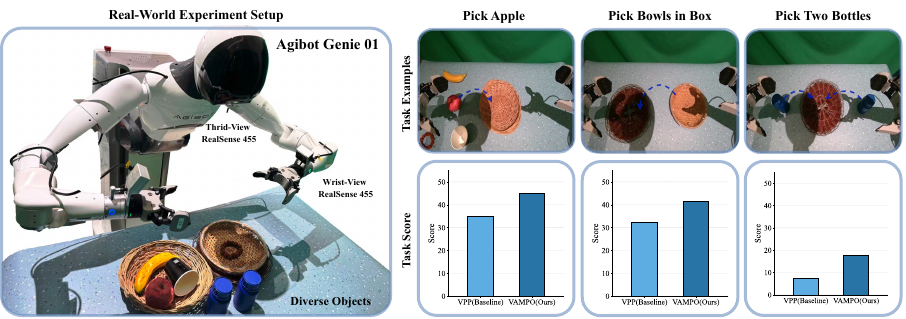}
    \caption{\textbf{Real-world evaluation across multiple task benchmarks.} The figure reports performance on three manipulation benchmarks---grasping in clutter, single-arm pick-and-place, and bimanual grasp-and-place---on the Agibot Genie 01 platform. \methodname (Ours) achieves the best performance across all tasks.}
    \label{fig:real}
\end{figure*}

\textbf{Result Analysis.} Figure~\ref{fig:vis_appendix} shows that \methodname{} consistently improves video generation quality across three distinct scenes and background conditions, highlighting the generalizability of our approach.
Figure~\ref{fig:real} further shows that \methodname{} consistently outperforms the base policy in real-world scenarios, demonstrating the effectiveness and robustness of the proposed method. Detailed task scoring criteria are provided in Appendix~\ref{app:Real-world Experiment Details}.

\begin{figure*}[t]  
    \centering
    \includegraphics[width=\textwidth, trim=0 0 0 0, clip]{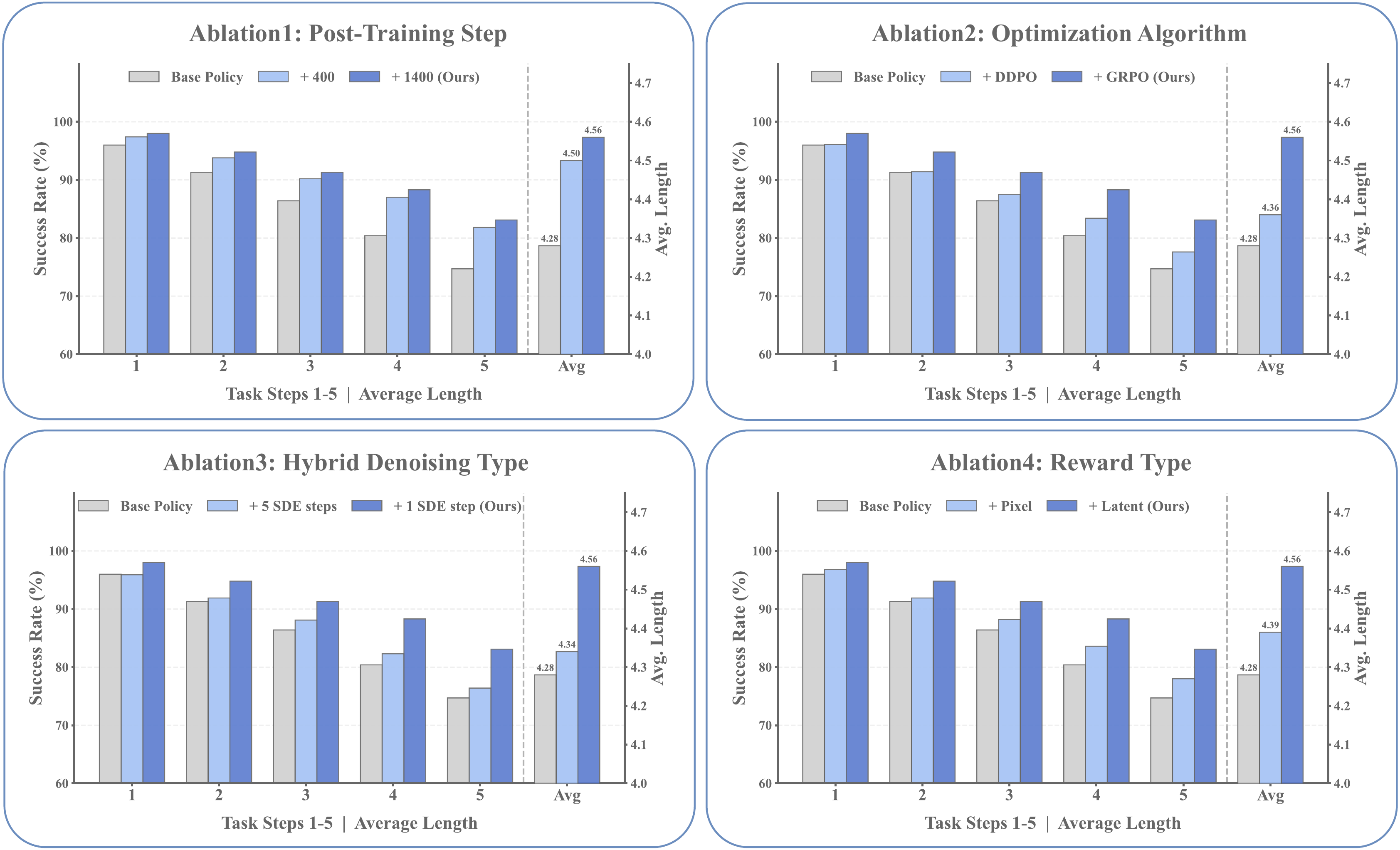}
    \caption{\textbf{Ablation studies on core components.} The figure visualizes the effects of different design choices on CALVIN ABC$\to$D: (a) post-training steps (GRPO iterations), (b) optimization algorithm (DDPO vs.\ GRPO), (c) hybrid denoising strategy (5-step vs.\ 1-step SDE), and (d) reward type (pixel-level vs.\ latent-level). Our final configuration (1400-step GRPO, 1-step SDE, latent-space reward) consistently achieves the best performance.}
    \label{fig:ablation}
\end{figure*}

\subsection{Components-wise Analysis }

Next, we perform an ablation study on the core components of our approach to evaluate the effectiveness of each design choice. All ablations are conducted in the CALVIN environment, with detailed numerical results provided in the Appendix.

\textbf{Post-Training Steps.} 
From Table~\ref{tab:ab0} and the corresponding trends in Figure~\ref{fig:ablation}, we observe that our method reaches a relatively strong performance within only 400 post-training steps. However, the overall training process is somewhat unstable, with occasional performance drops in the middle of training. Despite these fluctuations, our method consistently outperforms the base policy, and achieves its best performance at 1400 steps. This trend reflects the inherent stochasticity of reinforcement learning; importantly, the results at every evaluated checkpoint remain higher than the base policy, demonstrating the robustness and effectiveness of our approach.

\textbf{Optimization Algorithm.}
As shown in Table~\ref{tab:ab1} and Figure~\ref{fig:ablation}, GRPO consistently outperforms DDPO under the same base policy, yielding higher success rates across all horizons and a larger improvement in average trajectory length. We hypothesize that the within-group optimization and clipping scheme in GRPO lead to more stable and informative gradient updates, whereas DDPO is more sensitive to noisy rewards and can underutilize the diversity of candidate rollouts.

\textbf{Hybrid Denoising Type.}
In Table~\ref{tab:ab2} and Figure~\ref{fig:ablation}, we verify our motivation for applying SDE to the first step only. This is mainly because applying SDE to later steps can lead to reward hacking: the first-step output is fed directly to the policy, whereas perturbing later steps may encourage the model to exploit the reward in unintended ways.

\textbf{Reward Type.}
For reward computation, we consider two alternatives: computing the reward directly in the latent space, or decoding latents back to the image space and using a pixel-level reward. As shown in Table~\ref{tab:ab3} and Figure~\ref{fig:ablation}, the latent-space reward achieves better performance. We attribute this to the fact that pixel-level consistency does not necessarily imply more accurate visual dynamics modeling, whereas latent representations are more aligned with the underlying dynamics and task-relevant semantics.

\section{Related Work}

\noindent
\textbf{Video-Action Models.}
Video Action Model-based VLAs represent a more fundamentally novel paradigm than traditional VLM-based VLAs.
Recent works on Video Action Models \cite{hu2024video, zhang2025gevrm, pai2025mimic, kim2026cosmos, wang2025unified} have introduced a new direction for VLA modeling by explicitly incorporating future visual prediction into action generation. In contrast, conventional VLM-based VLAs \cite{gong2025carp,brohan2022rt,zitkovich2023rt,li2023vision,kim2024openvla,song2025reconvla}
typically learn a direct mapping from current observations and language instructions to actions, without explicitly modeling the underlying system dynamics.
This predictive structure enables the policy to reason over temporal evolution and action consequences, rather than relying solely on static representations.
Video Action Models (VAM) \cite{hu2024video, zhang2025gevrm, pai2025mimic, kim2026cosmos, wang2025unified, ye2026world} have recently emerged as a novel paradigm in the field of Language Action Model(VLA). Unlike traditional Vision-Language Models (VLM)-based VLAs \cite{gong2025carp,brohan2022rt,zitkovich2023rt,li2023vision,kim2024openvla,song2025reconvla}, which directly map observations and language to the action space, VAM predicts future observations based on language and current observations to guide action generation. Therefore, VAM provides a stronger dynamical prior for the action policy by predicting future observations. This approach helps improve the accuracy of action generation, particularly in capturing complex temporal dynamics and dependencies between actions \cite{hu2024video}. While VAM improves action generation by providing stronger dynamical priors, the quality of the generated video can still affect the accuracy of the actions. For example, training with likelihood surrogates inevitably introduces issues like misalignment, blurriness, and hallucinations, leading to potential inaccuracies in action generation. To address these issues, our work focuses on optimizing the video generation model with policy-relevant representations, enhancing dynamic feature extraction and further improving the accuracy and reliability of action generation.


\noindent
\textbf{Diffusion Models for Robot Control.}
Reinforcement learning has recently been integrated with diffusion models to improve generation quality and policy behaviors. DDPO \cite{black2023training} formulates denoising as a sequential decision process. GRPO-based extensions for generative models—including flow-based variants \cite{liu2025flow, xue2025dancegrpo}, hierarchical and structural variants \cite{ding2025treegrpo, li2025branchgrpo}, and generative alignment frameworks \cite{wang2025pref, zheng2025diffusionnft}—further explore combining RL’s long-term reward optimization with the expressive power of diffusion models.
Our work adapts similar principles to optimize video prediction models in latent space, explicitly treating prediction quality as an intermediate objective and strengthening the dynamical prior for downstream VLA policies. 

\section{Conclusion}
This paper introduced \methodname{}, a reward-based post-training framework for improving visual dynamics in video action models. By identifying the objective mismatch between likelihood-based generative training and action-centric deployment, we proposed to cast multi-step denoising as a sequential decision process and directly optimize the resulting future representations with rewards defined over expert visual dynamics. With an Euler Hybrid sampler and GRPO-based optimization, \methodname{} provides an effective and tractable way to improve precision-critical dynamics for downstream manipulation.
Our experiments across simulated and real-world settings show that better visual dynamics lead to better action generation and stronger generalization. More broadly, our results suggest that moving beyond likelihood-only training and toward direct optimization of control-relevant predictive signals may be an important direction for future video-based robot learning and VLA systems.
\clearpage


\bibliographystyle{assets/plainnat}
\bibliography{paper}

@inproceedings{roy2007effective,
  title={The effective rank: A measure of effective dimensionality},
  author={Roy, Olivier and Vetterli, Martin},
  booktitle={2007 15th European signal processing conference},
  pages={606--610},
  year={2007},
  organization={IEEE}
}

@article{ye2026world,
  title={World action models are zero-shot policies},
  author={Ye, Seonghyeon and Ge, Yunhao and Zheng, Kaiyuan and Gao, Shenyuan and Yu, Sihyun and Kurian, George and Indupuru, Suneel and Tan, You Liang and Zhu, Chuning and Xiang, Jiannan and others},
  journal={arXiv preprint arXiv:2602.15922},
  year={2026}
}

@article{hu2024video,
  title={Video prediction policy: A generalist robot policy with predictive visual representations},
  author={Hu, Yucheng and Guo, Yanjiang and Wang, Pengchao and Chen, Xiaoyu and Wang, Yen-Jen and Zhang, Jianke and Sreenath, Koushil and Lu, Chaochao and Chen, Jianyu},
  journal={arXiv preprint arXiv:2412.14803},
  year={2024}
}

@article{black2023training,
  title={Training diffusion models with reinforcement learning},
  author={Black, Kevin and Janner, Michael and Du, Yilun and Kostrikov, Ilya and Levine, Sergey},
  journal={arXiv preprint arXiv:2305.13301},
  year={2023}
}

@article{karras2022elucidating,
  title={Elucidating the design space of diffusion-based generative models},
  author={Karras, Tero and Aittala, Miika and Aila, Timo and Laine, Samuli},
  journal={Advances in neural information processing systems},
  volume={35},
  pages={26565--26577},
  year={2022}
}

@article{blattmann2023stable,
  title={Stable video diffusion: Scaling latent video diffusion models to large datasets},
  author={Blattmann, Andreas and Dockhorn, Tim and Kulal, Sumith and Mendelevitch, Daniel and Kilian, Maciej and Lorenz, Dominik and Levi, Yam and English, Zion and Voleti, Vikram and Letts, Adam and others},
  journal={arXiv preprint arXiv:2311.15127},
  year={2023}
}

@article{chi2025diffusion,
  title={Diffusion policy: Visuomotor policy learning via action diffusion},
  author={Chi, Cheng and Xu, Zhenjia and Feng, Siyuan and Cousineau, Eric and Du, Yilun and Burchfiel, Benjamin and Tedrake, Russ and Song, Shuran},
  journal={The International Journal of Robotics Research},
  volume={44},
  number={10-11},
  pages={1684--1704},
  year={2025},
  publisher={Sage Publications Sage UK: London, England}
}

@article{schulman2017proximal,
  title={Proximal policy optimization algorithms},
  author={Schulman, John and Wolski, Filip and Dhariwal, Prafulla and Radford, Alec and Klimov, Oleg},
  journal={arXiv preprint arXiv:1707.06347},
  year={2017}
}

@article{ho2020denoising,
  title={Denoising diffusion probabilistic models},
  author={Ho, Jonathan and Jain, Ajay and Abbeel, Pieter},
  journal={Advances in neural information processing systems},
  volume={33},
  pages={6840--6851},
  year={2020}
}

@article{song2020denoising,
  title={Denoising diffusion implicit models},
  author={Song, Jiaming and Meng, Chenlin and Ermon, Stefano},
  journal={arXiv preprint arXiv:2010.02502},
  year={2020}
}

@article{brohan2022rt,
  title={Rt-1: Robotics transformer for real-world control at scale},
  author={Brohan, Anthony and Brown, Noah and Carbajal, Justice and Chebotar, Yevgen and Dabis, Joseph and Finn, Chelsea and Gopalakrishnan, Keerthana and Hausman, Karol and Herzog, Alex and Hsu, Jasmine and others},
  journal={arXiv preprint arXiv:2212.06817},
  year={2022}
}

@inproceedings{zitkovich2023rt,
  title={Rt-2: Vision-language-action models transfer web knowledge to robotic control},
  author={Zitkovich, Brianna and Yu, Tianhe and Xu, Sichun and Xu, Peng and Xiao, Ted and Xia, Fei and Wu, Jialin and Wohlhart, Paul and Welker, Stefan and Wahid, Ayzaan and others},
  booktitle={Conference on Robot Learning},
  pages={2165--2183},
  year={2023},
  organization={PMLR}
}

@article{li2023vision,
  title={Vision-language foundation models as effective robot imitators},
  author={Li, Xinghang and Liu, Minghuan and Zhang, Hanbo and Yu, Cunjun and Xu, Jie and Wu, Hongtao and Cheang, Chilam and Jing, Ya and Zhang, Weinan and Liu, Huaping and others},
  journal={arXiv preprint arXiv:2311.01378},
  year={2023}
}

@inproceedings{gong2025carp,
  title={Carp: Visuomotor policy learning via coarse-to-fine autoregressive prediction},
  author={Gong, Zhefei and Ding, Pengxiang and Lyu, Shangke and Huang, Siteng and Sun, Mingyang and Zhao, Wei and Fan, Zhaoxin and Wang, Donglin},
  booktitle={Proceedings of the IEEE/CVF International Conference on Computer Vision},
  pages={13460--13470},
  year={2025}
}

@article{song2025reconvla,
  title={Reconvla: Reconstructive vision-language-action model as effective robot perceiver},
  author={Song, Wenxuan and Zhou, Ziyang and Zhao, Han and Chen, Jiayi and Ding, Pengxiang and Yan, Haodong and Huang, Yuxin and Tang, Feilong and Wang, Donglin and Li, Haoang},
  journal={arXiv preprint arXiv:2508.10333},
  year={2025}
}

@article{kim2024openvla,
  title={Openvla: An open-source vision-language-action model},
  author={Kim, Moo Jin and Pertsch, Karl and Karamcheti, Siddharth and Xiao, Ted and Balakrishna, Ashwin and Nair, Suraj and Rafailov, Rafael and Foster, Ethan and Lam, Grace and Sanketi, Pannag and others},
  journal={arXiv preprint arXiv:2406.09246},
  year={2024}
}

@article{mees2022calvin,
  title={Calvin: A benchmark for language-conditioned policy learning for long-horizon robot manipulation tasks},
  author={Mees, Oier and Hermann, Lukas and Rosete-Beas, Erick and Burgard, Wolfram},
  journal={IEEE Robotics and Automation Letters},
  volume={7},
  number={3},
  pages={7327--7334},
  year={2022},
  publisher={IEEE}
}

@article{song2020score,
  title={Score-based generative modeling through stochastic differential equations},
  author={Song, Yang and Sohl-Dickstein, Jascha and Kingma, Diederik P and Kumar, Abhishek and Ermon, Stefano and Poole, Ben},
  journal={arXiv preprint arXiv:2011.13456},
  year={2020}
}

@article{kingma2021variational,
  title={Variational diffusion models},
  author={Kingma, Diederik and Salimans, Tim and Poole, Ben and Ho, Jonathan},
  journal={Advances in neural information processing systems},
  volume={34},
  pages={21696--21707},
  year={2021}
}

@article{du2023learning,
  title={Learning universal policies via text-guided video generation},
  author={Du, Yilun and Yang, Sherry and Dai, Bo and Dai, Hanjun and Nachum, Ofir and Tenenbaum, Josh and Schuurmans, Dale and Abbeel, Pieter},
  journal={Advances in neural information processing systems},
  volume={36},
  pages={9156--9172},
  year={2023}
}

@article{zhang2025gevrm,
  title={Gevrm: Goal-expressive video generation model for robust visual manipulation},
  author={Zhang, Hongyin and Ding, Pengxiang and Lyu, Shangke and Peng, Ying and Wang, Donglin},
  journal={arXiv preprint arXiv:2502.09268},
  year={2025}
}

@article{pai2025mimic,
  title={mimic-video: Video-action models for generalizable robot control beyond vlas},
  author={Pai, Jonas and Achenbach, Liam and Montesinos, Victoriano and Forrai, Benedek and Mees, Oier and Nava, Elvis},
  journal={arXiv preprint arXiv:2512.15692},
  year={2025}
}

@article{kim2026cosmos,
  title={Cosmos policy: Fine-tuning video models for visuomotor control and planning},
  author={Kim, Moo Jin and Gao, Yihuai and Lin, Tsung-Yi and Lin, Yen-Chen and Ge, Yunhao and Lam, Grace and Liang, Percy and Song, Shuran and Liu, Ming-Yu and Finn, Chelsea and others},
  journal={arXiv preprint arXiv:2601.16163},
  year={2026}
}

@article{wang2025unified,
  title={Unified vision-language-action model},
  author={Wang, Yuqi and Li, Xinghang and Wang, Wenxuan and Zhang, Junbo and Li, Yingyan and Chen, Yuntao and Wang, Xinlong and Zhang, Zhaoxiang},
  journal={arXiv preprint arXiv:2506.19850},
  year={2025}
}

@article{kim2025fine,
  title={Fine-tuning vision-language-action models: Optimizing speed and success},
  author={Kim, Moo Jin and Finn, Chelsea and Liang, Percy},
  journal={arXiv preprint arXiv:2502.19645},
  year={2025}
}

@article{tian2024predictive,
  title={Predictive inverse dynamics models are scalable learners for robotic manipulation},
  author={Tian, Yang and Yang, Sizhe and Zeng, Jia and Wang, Ping and Lin, Dahua and Dong, Hao and Pang, Jiangmiao},
  journal={arXiv preprint arXiv:2412.15109},
  year={2024}
}

@article{black2024pi_0,
  title={$\pi_0$: A Vision-Language-Action Flow Model for General Robot Control},
  author={Black, Kevin and Brown, Noah and Driess, Danny and Esmail, Adnan and Equi, Michael and Finn, Chelsea and Fusai, Niccolo and Groom, Lachy and Hausman, Karol and Ichter, Brian and others},
  journal={arXiv preprint arXiv:2410.24164},
  year={2024}
}

@article{wu2023unleashing,
  title={Unleashing large-scale video generative pre-training for visual robot manipulation},
  author={Wu, Hongtao and Jing, Ya and Cheang, Chilam and Chen, Guangzeng and Xu, Jiafeng and Li, Xinghang and Liu, Minghuan and Li, Hang and Kong, Tao},
  journal={arXiv preprint arXiv:2312.13139},
  year={2023}
}

@article{wen2024vidman,
  title={Vidman: Exploiting implicit dynamics from video diffusion model for effective robot manipulation},
  author={Wen, Youpeng and Lin, Junfan and Zhu, Yi and Han, Jianhua and Xu, Hang and Zhao, Shen and Liang, Xiaodan},
  journal={Advances in Neural Information Processing Systems},
  volume={37},
  pages={41051--41075},
  year={2024}
}

@article{intelligence2025pi_,
  title={$\pi_{0.5}$: a Vision-Language-Action Model with Open-World Generalization},
  author={Intelligence, Physical and Black, Kevin and Brown, Noah and Darpinian, James and Dhabalia, Karan and Driess, Danny and Esmail, Adnan and Equi, Michael and Finn, Chelsea and Fusai, Niccolo and others},
  journal={arXiv preprint arXiv:2504.16054},
  year={2025}
}

@article{bu2024closed,
  title={Closed-loop visuomotor control with generative expectation for robotic manipulation},
  author={Bu, Qingwen and Zeng, Jia and Chen, Li and Yang, Yanchao and Zhou, Guyue and Yan, Junchi and Luo, Ping and Cui, Heming and Ma, Yi and Li, Hongyang},
  journal={Advances in Neural Information Processing Systems},
  volume={37},
  pages={139002--139029},
  year={2024}
}

@article{black2023zero,
  title={Zero-shot robotic manipulation with pretrained image-editing diffusion models},
  author={Black, Kevin and Nakamoto, Mitsuhiko and Atreya, Pranav and Walke, Homer and Finn, Chelsea and Kumar, Aviral and Levine, Sergey},
  journal={arXiv preprint arXiv:2310.10639},
  year={2023}
}

@article{fan2025long,
  title={Long-vla: Unleashing long-horizon capability of vision language action model for robot manipulation},
  author={Fan, Yiguo and Ding, Pengxiang and Bai, Shuanghao and Tong, Xinyang and Zhu, Yuyang and Lu, Hongchao and Dai, Fengqi and Zhao, Wei and Liu, Yang and Huang, Siteng and others},
  journal={arXiv preprint arXiv:2508.19958},
  year={2025}
}

@article{liu2025trivla,
  title={Trivla: A triple-system-based unified vision-language-action model for general robot control},
  author={Liu, Zhenyang and Gu, Yongchong and Zheng, Sixiao and Xue, Xiangyang and Fu, Yanwei},
  journal={arXiv e-prints},
  pages={arXiv--2507},
  year={2025}
}

@article{liu2025flow,
  title={Flow-grpo: Training flow matching models via online rl},
  author={Liu, Jie and Liu, Gongye and Liang, Jiajun and Li, Yangguang and Liu, Jiaheng and Wang, Xintao and Wan, Pengfei and Zhang, Di and Ouyang, Wanli},
  journal={arXiv preprint arXiv:2505.05470},
  year={2025}
}

@article{xue2025dancegrpo,
  title={Dancegrpo: Unleashing grpo on visual generation},
  author={Xue, Zeyue and Wu, Jie and Gao, Yu and Kong, Fangyuan and Zhu, Lingting and Chen, Mengzhao and Liu, Zhiheng and Liu, Wei and Guo, Qiushan and Huang, Weilin and others},
  journal={arXiv preprint arXiv:2505.07818},
  year={2025}
}

@article{ding2025treegrpo,
  title={TreeGRPO: Tree-Advantage GRPO for Online RL Post-Training of Diffusion Models},
  author={Ding, Zheng and Ye, Weirui},
  journal={arXiv preprint arXiv:2512.08153},
  year={2025}
}

@article{li2025branchgrpo,
  title={Branchgrpo: Stable and efficient grpo with structured branching in diffusion models},
  author={Li, Yuming and Wang, Yikai and Zhu, Yuying and Zhao, Zhongyu and Lu, Ming and She, Qi and Zhang, Shanghang},
  journal={arXiv preprint arXiv:2509.06040},
  year={2025}
}

@article{wang2025pref,
  title={Pref-grpo: Pairwise preference reward-based grpo for stable text-to-image reinforcement learning},
  author={Wang, Yibin and Li, Zhimin and Zang, Yuhang and Zhou, Yujie and Bu, Jiazi and Wang, Chunyu and Lu, Qinglin and Jin, Cheng and Wang, Jiaqi},
  journal={arXiv preprint arXiv:2508.20751},
  year={2025}
}

@article{zheng2025diffusionnft,
  title={Diffusionnft: Online diffusion reinforcement with forward process},
  author={Zheng, Kaiwen and Chen, Huayu and Ye, Haotian and Wang, Haoxiang and Zhang, Qinsheng and Jiang, Kai and Su, Hang and Ermon, Stefano and Zhu, Jun and Liu, Ming-Yu},
  journal={arXiv preprint arXiv:2509.16117},
  year={2025}
}

\clearpage
\newpage
\onecolumn
\beginappendix
\renewcommand{\thefigure}{S\arabic{figure}}
\renewcommand{\thetable}{S\arabic{table}}
\setcounter{figure}{0}
\setcounter{table}{0}

\label{sec:appendix}

\section{Training Details}
We perform post-training on a VPM initialized from the pretrained checkpoint provided by VPP. The training data consists of videos from the CALVIN ABC dataset, which contains diverse long-horizon manipulation trajectories. In this stage, the VPM is optimized using reinforcement learning for 1.5k training steps with 64 NVIDIA H20 GPUs. Training is conducted in a distributed data-parallel manner. After VPM post-training, we train the AGM on the CALVIN ABC dataset for around 10 epochs, using 8 NVIDIA H20 GPUs. Unless otherwise specified, all evaluation is implemented  on NVIDIA RTX 5880 GPUs.

\begin{table}[htbp]
  \centering
  \small
  \begin{tabular}{llc}
    \toprule
    Type & Name & Parameters \\
    \midrule
    \multirow{2}{*}{Prediction}
      & Video length & 16 \\
      & Action shape & $10 \times 7$ \\
    \midrule
    \multirow{2}{*}{TVP}
      & Language shape & $20 \times 512$ \\
      & Image shape & $256 \times 256$ \\
    \midrule
    \multirow{5}{*}{Video Former}
      & Token shape & $16 \times 14 \times 384$ \\
      & Input dim & 1280 \\
      & Latent dim & 512 \\
      & Num heads & 8 \\
      & Num layers & 6 \\
    \midrule
    \multirow{6}{*}{Diffusion Transformer}
      & Latent dim & 384 \\
      & Condition shape & $225 \times 384$ \\
      & Num heads & 8 \\
      & Encoder layers & 4 \\
      & Decoder layers & 4 \\
      & Sampling steps & 10 \\
    \bottomrule
  \end{tabular}
    \caption{\textbf{Model Configuration Parameters.} Detailed architectural specifications for our model, including tensor shapes for the Video Former and Diffusion Transformer components}
  \label{tab:config}
\end{table}

\begin{table}[htbp]
  \centering
  \small
  \begin{tabular}{llc}
    \toprule
    Category & Parameter & Value \\
    \midrule
    \multirow{3}{*}{Optimization} 
      & Group size ($G$) & 8 \\
      & PPO clipping parameter ($\epsilon_{\text{clip}}$) & 0.2 \\
      & Batch size & 8 \\
    \midrule
    \multirow{2}{*}{Learning Rate} 
      & Fine-tuning & $1 \times 10^{-4}$ \\
      & Post-training & $1 \times 10^{-6}$ \\
    \midrule
    \multirow{2}{*}{Reward Weights} 
      & $L_1$ reward & 1.0 \\
      & Cosine reward & 1.0 \\
    \bottomrule
  \end{tabular}
  \caption{Summary of hyperparameters and optimization settings.}
  \label{tab:hyperparameters}
\end{table}

\section{Real-world Experiment Details}
\label{app:Real-world Experiment Details}

\medskip
\noindent\textbf{Real-World Tasks.}
We design three representative tabletop manipulation tasks to evaluate semantic understanding, visual perception, precise control, and dual-arm coordination.

\begin{center}
    \nopagebreak
    \makeatletter
    \def\@captype{figure}
    \makeatother
    \includegraphics[
        width=\linewidth,
        trim=0cm 0cm 0cm 0cm, 
        clip
    ]{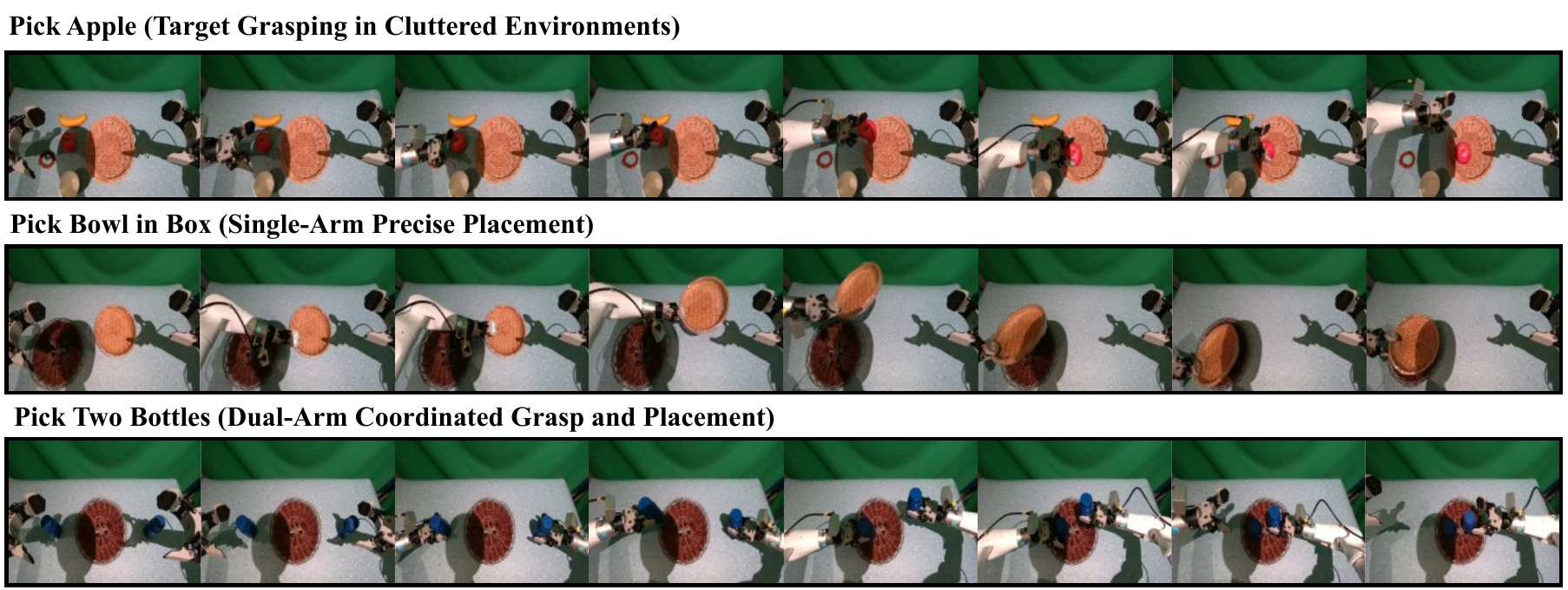}
    \caption{Real-world robot execution examples.} 
    \label{fig:real_eg}
\end{center}

\medskip
\noindent\textbf{(1) Target Grasping in Cluttered Environments (\texttt{pick\_apple}).}
The robot must grasp a specified target object (apple) from a cluttered tabletop containing irrelevant objects, testing robustness under visual distraction and accurate localization. Scoring: target grasp success (50), successful placement into plate (50), total 100.

\medskip
\noindent\textbf{(2) Single-Arm Precise Placement (\texttt{pick\_bowl\_in\_box}).}
The robot picks up a bowl from one side of the table and places it onto a centrally located plate with required positional and orientation precision. This task evaluates spatial reasoning, grasp stability, and fine-grained placement control. Scoring: target grasp success (50), successful placement onto plate (50), total 100.

\medskip
\noindent\textbf{(3) Dual-Arm Coordinated Grasp and Placement (\texttt{pick\_two\_bottles}).}
Two bottles are placed on the table. The robot simultaneously grasps both bottles using dual arms and places them onto a central plate. This task evaluates multi-object perception, temporal and spatial coordination, synchronized control, and complex sequence planning. Scoring: left-arm grasp (25), right-arm grasp (25), left-arm placement (25), right-arm placement (25), total 100.

\section{Ablations}
\begin{table}[!htbp]
    \centering
    \small
	\setlength{\tabcolsep}{2pt}
	\caption{\textbf{Ablation on reward type: pixel-level vs.\ latent-level reward on CALVIN ABC$\to$D.} Latent-space reward (Ours) achieves the best performance.}
    \label{tab:ab3}
	\begin{tabular}{cc|ccccc|c}
		\toprule
		\multicolumn{2}{c|}{} & \multicolumn{5}{c|}{Task completed in a row $\uparrow$} & Avg.\ Len $\uparrow$ \\
		\cmidrule(lr){3-7}
		\multicolumn{2}{c|}{} & 1 & 2 & 3 & 4 & 5 & \\
		\midrule
		\multirow{1}{*}{} & Base Policy & 96.0 & 91.3 & 86.4 & 80.4 & 74.7 & 4.28 \\
		\midrule
		\multirow{1}{*}{+} & Pixel & 96.8 & 91.9 & 88.2 & 83.6 & 78.0 & 4.39 \textcolor{green!55!black}{(+\,0.08)} \\
		\midrule
		\rowcolor[rgb]{.96,.96,.96}\multirow{1}{*}{+} & \textbf{Latent (Ours)} & \textbf{98.0} & \textbf{94.8} & \textbf{91.3} & \textbf{88.3} & \textbf{83.1} & \textbf{4.56} \textcolor{green!55!black}{(+\,0.28)} \\
		\bottomrule
	\end{tabular}%
\end{table}%

\begin{table}[!htbp]
    \centering
    \small
	\setlength{\tabcolsep}{2pt}
	\caption{\textbf{Ablation on hybrid denoising: SDE at 5 steps vs.\ 1 step.} Applying SDE only at the first step (Ours) yields the best performance and mitigates reward hacking.}
    \label{tab:ab2}
	\begin{tabular}{cc|ccccc|c}
		\toprule
		\multicolumn{2}{c|}{} & \multicolumn{5}{c|}{Task completed in a row $\uparrow$} & Avg.\ Len $\uparrow$ \\
		\cmidrule(lr){3-7}
		\multicolumn{2}{c|}{} & 1 & 2 & 3 & 4 & 5 & \\
		\midrule
		\multirow{1}{*}{} & \textbf{Base Policy} & 96.0 & 91.3 & 86.4 & 80.4 & 74.7 & 4.28 \\
		\midrule
		\multirow{1}{*}{+} & 5-step SDE & 95.9 & 91.9 & 88.1 & 82.3 & 76.4 & 4.34 \textcolor{green!55!black}{(+\,0.06)} \\
		\midrule
		\rowcolor[rgb]{.96,.96,.96}\multirow{1}{*}{+} & \textbf{1-step SDE (Ours)} & \textbf{98.0} & \textbf{94.8} & \textbf{91.3} & \textbf{88.3} & \textbf{83.1} & \textbf{4.56} \textcolor{green!55!black}{(+\,0.28)} \\
		\bottomrule
	\end{tabular}%
\end{table}%

\begin{table}[!htbp]
    \centering
    \small
	\setlength{\tabcolsep}{2pt}
	\caption{\textbf{Ablation on post-training steps (CALVIN ABC$\to$D).} Success rates and average length for different GRPO step counts. Best performance is achieved at 1400 steps; best values are in \textbf{bold}.}
    \label{tab:ab0}
	\begin{tabular}{cc|ccccc|c}
		\toprule
		\multicolumn{2}{c|}{} & \multicolumn{5}{c|}{Task completed in a row $\uparrow$} & Avg.\ Len $\uparrow$ \\
		\cmidrule(lr){3-7}
		\multicolumn{2}{c|}{} & 1 & 2 & 3 & 4 & 5 & \\
		\midrule
		\multirow{1}{*}{} & \textbf{Base Policy} & 96.0 & 91.3 & 86.4 & 80.4 & 74.7 & 4.28 \\
		\midrule
		\multirow{1}{*}{+} & 400 & 97.4 & 93.8 & 90.2 & 87.0 & 81.8 & 4.50 \textcolor{green!55!black}{(+\,0.22)} \\
		\multirow{1}{*}{+} & 600 & 95.6 & 90.6 & 85.6 & 81.8 & 77.2 & 4.31 \textcolor{green!55!black}{(+\,0.03)} \\
		\multirow{1}{*}{+} & 1000 & 96.5 & 92.8 & 89.0 & 84.9 & 78.2 & 4.42 \textcolor{green!55!black}{(+\,0.14)} \\
		\midrule
		\rowcolor[rgb]{.96,.96,.96}\multirow{1}{*}{+} & \textbf{1400 (Ours)} & \textbf{98.0} & \textbf{94.8} & \textbf{91.3} & \textbf{88.3} & \textbf{83.1} & \textbf{4.56} \textcolor{green!55!black}{(+\,0.28)} \\
		\bottomrule
	\end{tabular}%
\end{table}%

\begin{table}[!htbp]
    \centering
    \small
	\setlength{\tabcolsep}{2pt}
	\caption{\textbf{Ablation on optimization algorithm.} Comparison of DDPO and GRPO on CALVIN ABC$\to$D. GRPO (Ours) achieves the best performance.}
    \label{tab:ab1}
	\begin{tabular}{cc|ccccc|c}
		\toprule
		\multicolumn{2}{c|}{} & \multicolumn{5}{c|}{Task completed in a row $\uparrow$} & Avg.\ Len $\uparrow$ \\
		\cmidrule(lr){3-7}
		\multicolumn{2}{c|}{} & 1 & 2 & 3 & 4 & 5 & \\
		\midrule
		\multirow{1}{*}{} & \textbf{Base Policy} & 96.0 & 91.3 & 86.4 & 80.4 & 74.7 & 4.28 \\
		\midrule
		\multirow{1}{*}{+} & DDPO & 96.1 & 91.4 & 87.5 & 83.4 & 77.6 & 4.36 \textcolor{green!55!black}{(+\,0.08)} \\
		\midrule
		\rowcolor[rgb]{.96,.96,.96}\multirow{1}{*}{+} & \textbf{GRPO (Ours)} & \textbf{98.0} & \textbf{94.8} & \textbf{91.3} & \textbf{88.3} & \textbf{83.1} & \textbf{4.56} \textcolor{green!55!black}{(+\,0.28)} \\
		\bottomrule
	\end{tabular}%
\end{table}%




\section{Visualization}

\begin{center}
    \nopagebreak
    \makeatletter
    \def\@captype{figure}
    \makeatother
    \includegraphics[
        width=\linewidth,
        trim=0cm 1cm 0cm 1cm,
        clip
    ]{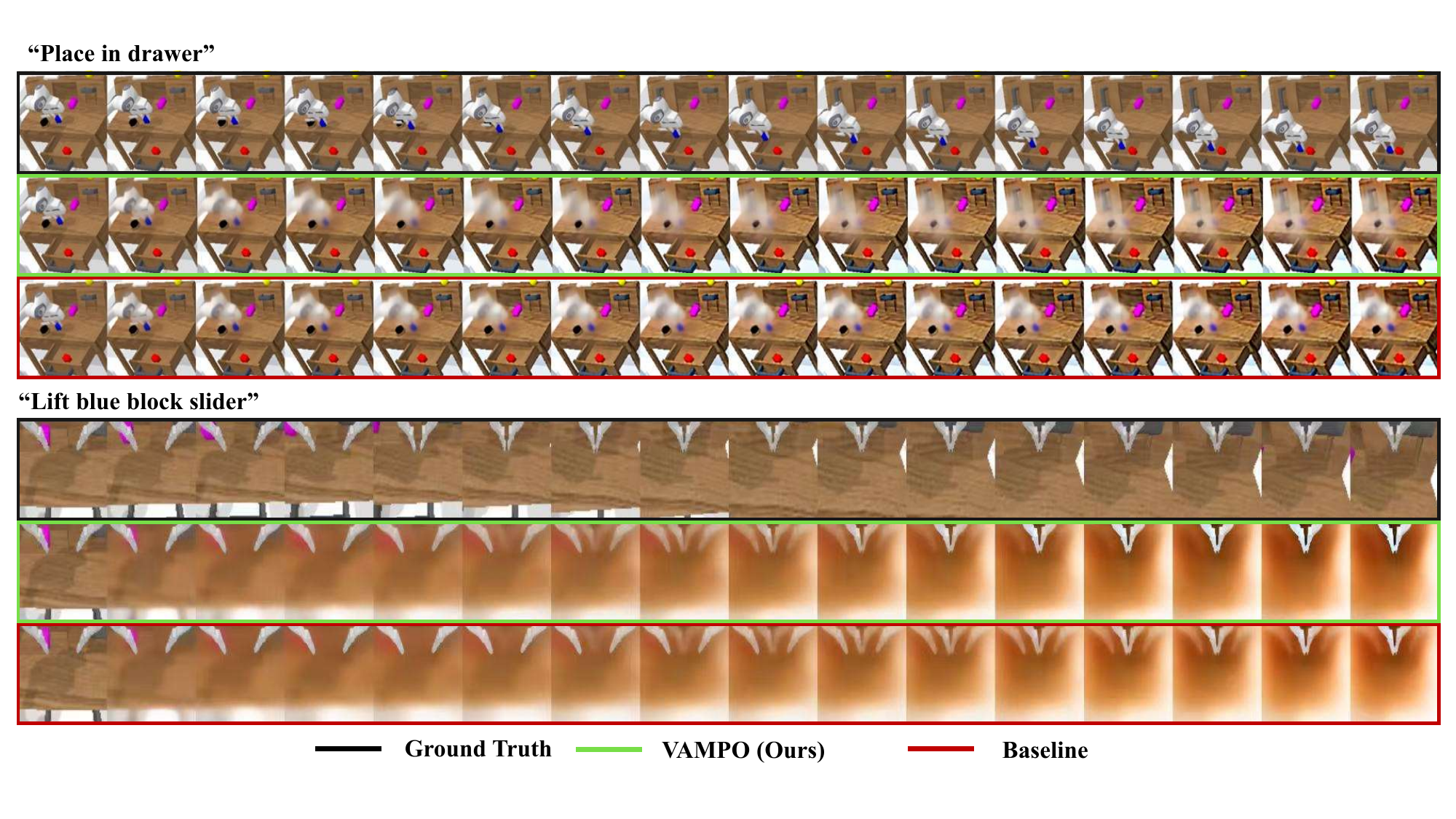}
    \caption{\textbf{Visualization of One-step Direct Prediction.} As illustrated, our model accurately captures the underlying visual dynamics of the robot's motion during the prediction process. Although the textures and fine-grained details are not perfectly precise, the consistent trajectory and physical interactions demonstrate the model's effectiveness}
    \label{fig:1step}
\end{center}

\end{document}